\documentclass{article}

\PassOptionsToPackage{numbers, compress}{natbib}

\usepackage[final]{neurips_data_2021}

\clearpage{}

\usepackage[utf8]{inputenc} \usepackage[T1]{fontenc} 

\usepackage[dvipsnames]{xcolor}

\usepackage{amsfonts}
\usepackage{nicefrac}       

\usepackage{booktabs} 
\usepackage{multirow}
\usepackage{multicol}

\usepackage[hidelinks]{hyperref}
\usepackage{url}            

\usepackage{csquotes}

\usepackage{pifont}
\usepackage{textcomp}

\usepackage{standalone}
\usepackage{caption, subcaption}

\usepackage[normalem]{ulem}

\usepackage[acronym,nohypertypes={acronym,notation}]{glossaries}

\usepackage{enumitem}

\usepackage{microtype}      

\usepackage[nolist,nohyperlinks]{acronym}

\newcommand{\cf}{cf.}
\newcommand{\etal}{et~al.}
\newcommand{\eg}{e.g.}
\newcommand{\ie}{i.e.}

\newcommand{\toolstyling}[1]{#1}
\newcommand{\rawnet}{\toolstyling{RawNet2}}
\newcommand{\PWG}{\toolstyling{PWG}}
\newcommand{\MBMel}{\toolstyling{MB-MelGAN}}
\newcommand{\FBMel}{\toolstyling{FB-MelGAN}}
\newcommand{\Mel}{\toolstyling{MelGAN}}
\newcommand{\hifi}{\toolstyling{HiFi-GAN}}
\newcommand{\wav}{\toolstyling{WaveGlow}}

\newcommand{\FTTS}{\toolstyling{TTS}}

\newcommand{\datasetstyling}[1]{\textsc{#1}}
\newcommand{\lj}{\datasetstyling{LJSpeech}}
\newcommand{\jsut}{\datasetstyling{JSUT}}

\newcommand{\gray}[1]{\textcolor{gray}{#1}}

\clearpage{}

\author{Joel Frank \\
  Ruhr University Bochum\\
  Horst G\"ortz Institute for IT-Security \\
  \href{mailto:joel.frank@rub.de}{\texttt{joel.frank@rub.de}} \\
\And
  Lea Sch{\"o}nherr \\
  Ruhr University Bochum\\
  Horst G\"ortz Institute for IT-Security \\
  \href{mailto:lea.schoenherr@rub.de}{\texttt{lea.schoenherr@rub.de}} \\
}

\title{WaveFake: A Data Set to Facilitate \\Audio Deepfake Detection}

\begin{document}

\begin{acronym}
    \acro{DFT}{\emph{Discrete Fourier Transform}}
    \acro{EER}{\emph{Equal Error Rate}}
    \acro{TTS}{\emph{Text-To-Speech}}
    \acro{GMM}{\emph{Gaussian Mixture Model}}
    \acro{MFCC}{\emph{Mel Frequency Cepstral Coefficients}}
    \acro{LFCC}{\emph{Linear Frequency Cepstral Coefficients}}
    \acro{DCT}{\emph{Discrete Cosine Transform}}
    \acro{STFT}{\emph{Short-Time Fourier Transform}}
    \acro{DNN}{\emph{Deep Neural Network}}
    \acro{GAN}{\emph{Generative Adversarial Network}}
    \acro{ROC}{\emph{Receiver Operating Characteristic}}
    \acro{DFDC}{\emph{Deepfake Detection Challenge}}
    \acro{EM}{\emph{Expectation Maximization}}
\end{acronym}

\newcommand{\tts}{\ac{TTS}}
\newcommand{\gmm}{\ac{GMM}}
\newcommand{\mfcc}{\ac{MFCC}}
\newcommand{\lfcc}{\ac{LFCC}}
\newcommand{\dct}{\ac{DCT}}
\newcommand{\stft}{\ac{STFT}}
\newcommand{\dnn}{\ac{DNN}}
\newcommand{\gan}{\ac{GAN}}

\acrodefplural{GMM}[GMMs]{Gaussian Mixture Models}
\newcommand{\gmms}{\acp{GMM}}

\acrodefplural{DNN}[DNNs]{Deep Neural Networks}
\newcommand{\dnns}{\acp{DNN}}

\acrodefplural{GAN}[GANs]{Generative Adversarial Networks}
\newcommand{\gans}{\acp{GAN}} \maketitle

\begin{abstract}
Deep generative modeling has the potential to cause significant harm to society.
Recognizing this threat, a magnitude of research into detecting so-called ``Deepfakes'' has emerged.
This research most often focuses on the image domain, while studies exploring generated audio signals have--- so far ---been neglected.
In this paper, we aim to narrow this gap.
We present a novel data set, for which we collected ten sample sets from six different network architectures, spanning two languages.
We analyze the frequency statistics comprehensively, discovering subtle differences between the architectures, specifically among the higher frequencies.
Additionally, to facilitate further development of detection methods, we implemented three different classifiers adopted from the signal processing community to give practitioners a baseline to compare against.
In a first evaluation, we already discovered significant trade-offs between the different approaches.
Neural network-based approaches performed better on average, but more traditional models proved to be more robust.
\end{abstract}

\section{Introduction}
\label{intro}
\$243,000 were lost when criminals used a generated voice recording to impersonate the CEO of a UK company~\citep{stupp2019ceo}.
This is just one of several reports where current state-of-the-art generative modeling was used in harmful ways.
Other examples include: impersonation attempts~\citep{fran2020listen}, influencing opposition movements~\citep{dfrlab2021navalny}, being used to justify military actions~\citep{hao2019deepfake,mwai2021tigray}, or online harassment~\citep{burgess2020telegram}.
While there is a multitude of beneficial use cases, for example, enhancing data sets for medical diagnostics~\cite{esteban2017real, frid2018gan} or designing DNA to optimize protein bindings~\cite{killoran2017generating}, finding effective ways to detect fraudulent usage is of utmost importance to society.

Detection in the image domain has received tremendous attention~\citep{marra2018detection,mo2018fake, yu2019attributing,tariq2019gan,wang2019cnn,mccloskey2018detecting,nataraj2019detecting,marra2019gans,durall2020watch,frank2020leveraging}.
However, the audio domain is severely lacking.
In this paper, we aim at closing this gap.
We start by reviewing standard signal processing techniques used for analyzing audio signals.
For example, we give an introduction to spectrograms, which are commonly used as an intermediate representation for generative models~\citep{kumar2019melgan,prenger2019waveglow,yamamoto2020pwg,yang2021mbmelgan}.
Additionally, we provide a survey of current state-of-the-art generative models.

Our main contribution is a novel data set.
We collected ten sample sets from six different network architectures across two languages.
This paper focuses on analyzing samples that resemble (\ie, recreate) the training distributions.
This allows for one-to-one comparisons of audio clips between the different architectures.
In this comparison, we find subtle differences between the generators.
We also expect good performing classifiers to transfer well to other contexts since recreating the training distribution should yield the most quality samples.
We test this hypothesis by also generating completely novel phrases.

Finally, we implement three classifiers, which we adopted from best practices in the signal processing community~\citep{sahidullah2015comparison,tak2021rawnet2}, to give future researchers a baseline to compare against~\footnote{Our code and pretrained models can be found at~\href{https://github.com/RUB-SysSec/WaveFake}{\texttt{github.com/RUB-SysSec/WaveFake}}}.
In a first evaluation we already discovered trade-offs between the different approaches.
Neural networks performed better overall, but proved to be susceptible to changing settings.
Finally, we implemented BlurIG~\citep{xu2020attribution} a popular attribution method/package, so practitioners can inspect their predictions when building on our results.

We summarize our main contributions as follows:
\begin{itemize}[noitemsep,topsep=0pt,parsep=0pt,partopsep=0pt]
    \item
    A novel data set consisting of samples from several SOTA network architectures. Additionally, we perform a comprehensive analysis of this data set and find subtle differences between the different architectures.
    \item 
    An implementation of two baseline models for future researchers to compare against. These models were evaluated in three different settings and we provide a popular attribution method to inspect the prediction.
\end{itemize} \section{Background}
\label{background}

In this section, we provide an introduction to standard techniques used for analyzing speech audio signals.
For additional material on the topic, the interested reader is referred to the excellent books by ~\citet{rabiner2016theory} or~\citet{quatieri2006discrete}.
Additionally, we provide a survey on current SOTA generative models and summarize related work.

\subsection{Analyzing speech signals}
\label{background:features}

\paragraph{(Mel) spectrograms:}

A spectrogram is a visual representation of the frequency information of a signal over time (\cf{} Section~\ref{dataset}, Figure~\ref{fig:specs_reduced} for an example).
To calculate a spectrogram for an audio signal, we first divide the waveform into \emph{frames} (\eg, $20$\,ms) with an overlap (\eg, $10$\,ms) between two adjacent frames.
We then apply a window function to avoid spectral leakage~\footnote{Energies from one frequency leak into other frequency bins.}.
These functions (\eg, Hamming, Hann, Blackman window) are a trade-off between frequency resolution and spectral leakage. Their choice depends on the task and the signal properties, \cf{} \citet{prabhu2014window} for a detailed overview. 
The frames are then transformed individually using the \ac{DFT} to obtain a representation in the frequency domain $X(t,k)$.
Where $t = 1, \dots, T$ is the frame index of the signal and $k = 0,\dots, K-1$ are the \ac{DFT} coefficients.
Finally, we calculate the squared magnitude $|X(t,k)|^2$ of the complex-valued signal to obtain our final representation---the spectrogram.

A commonly used variant is the so-called Mel spectrogram.
It is motivated by studies that have shown that humans do not perceive frequencies on a linear scale.
In particular, they can detect differences in lower frequencies with a higher resolution when compared to higher frequencies~\citep{zwicker-07-psychoacoustics}.
The Mel scale is an empirically determined non-linear transformation that approximates this relationship:
\begin{equation}
f_{\text{mel}} = 2595 \cdot \text{log}_{10}\bigg(1+\frac{f}{700}\bigg),
\end{equation}
where $f$ is the frequency in Hz and $f_{\text{mel}}$ the  Mel-scaled frequency.
To obtain the Mel spectrogram, we apply an ensemble of $S$ triangular filters~$H_{\mathrm{mel}}$ (we provide a visual representation in Section~8 of the supplementary material).
These filters have a linear distance between the triangle mid frequencies in the Mel scale, which results in a logarithmic increasing distance of the frequencies in the frequency domain
\begin{equation}
X_{\mathrm{mel}}(t,s) = \sum_{k=0}^{K-1}|X(t,k)|H_{\mathrm{mel}}(s,k) \quad \forall \, s = 1,\dots, S,
\end{equation}
which gives us the final Mel spectrogram.
Based on it, we can compute a common feature representation for audio analysis.

\paragraph{Mel Frequency Cepstral Coefficients:}
\ac{MFCC} are derived from a Mel-scaled spectrogram by applying a \ac{DCT} to the logarithm of the Mel-filtered signal 
\begin{equation}
c(t, r) = \sum_{s=0}^{S-1}\log\big[X_{\mathrm{mel}}(t,s)\big] \cdot \text{cos}\bigg[\frac{\pi \cdot r\cdot(s+0.5)}{S}\bigg] \quad \forall \, r = 0,\dots, R-1,
\end{equation}
where $R$ is the number of DCT coefficients. 

\paragraph{Linear Frequency Cepstral Coefficients:}
We can also calculate \ac{LFCC}.
As the name suggests, these coefficients are derived by applying a linear filterbank (instead of a Mel filterbank) to the signal's spectrogram.
This results in retaining more high-frequency information.
Except for replacing the filter bank, all other steps remain the same as for \ac{MFCC} features.

\paragraph{(Double) Delta features:}\acp{MFCC} and \acp{LFCC} are often augmented by their first and second derivatives to represent the temporal structure of the input.
These are referred to as delta and double delta features, respectively.
In practice, these are often calculated by central difference approximation via
\begin{equation}
d(t) = \frac{\sum_{n=1}^N n\cdot\big[c\,(t+n)-c\,(t-n)\big]}{2\cdot\sum_{n=1}^N n^2} \quad \forall \, t = 0,\dots, T-1,
\end{equation}
where $d(t)$ is the delta at time $t$ and $N$ is a user-defined window length for computing the delta, and $c$ is either the \acp{MFCC}/\acp{LFCC} or the delta features (when calculating the double delta features).

\subsection{Text-to-speech (TTS)}
\label{background:tts}

\begin{figure}[t!]
    \centering
    \includegraphics[width=\textwidth]{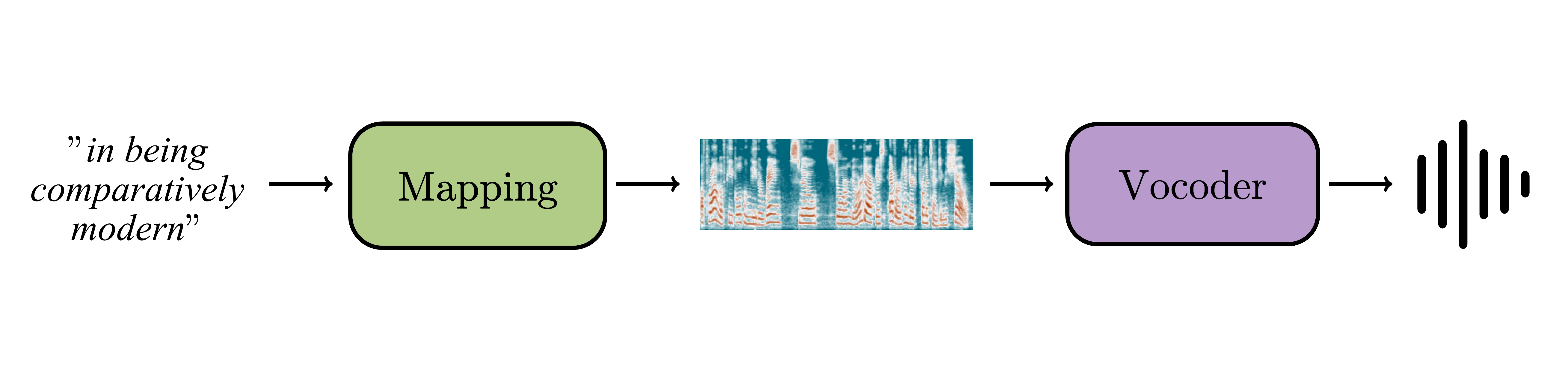}
    \vspace{-4em}
    \caption{
        \textbf{A typical TTS pipeline.}
        One model takes a textual prompt with the desired audio transcription (we call it the ``mapping'' model) and outputs an intermediate representation, for example Mel spectrograms.
        This intermediate representation is then fed to a second model (the ``vocoder'') to obtain the final raw audio.
}
    \label{fig:tts-pipeline}
\end{figure}

In this Section, we want to give a broad overview of different research directions for \tts{} models.
Due to the rapid developments of the field, this is a non-exhaustive list.

While there has been some research into end-to-end models~\citep{donahue2020end,taigman2017voiceloop}, typical \tts{} models consist of a two-stage approach, represented in Figure~\ref{fig:tts-pipeline}.
First, we enter the text sequence which we want to generate.
This sequence gets mapped by some model (or feature extraction method) to a low-dimensional intermediate representation, often linguistic features~\citep{arik2017deep} or Mel spectrograms~\citep{oord2016wavenet}.
Second, we use an additional model (often called vocoder) to map this intermediate representation to raw audio.
We focus on the literature on vocoders since it directly connects to our work.

Today, vocoders are typically implemented by \dnns{}.
The first \dnn{}~\citep{ze2013statistical,fan2014tts} approaches adopted the parametric vocoders of earlier HMM-based models~\citep{zen2009statistical,tokuda2000speech,yoshimura1999simultaneous}.
Here the \dnn{} was used to predict the statistics of a given time frame, which are then used in traditional speech parameter generation algorithms~\citep{tokuda2000speech}.
Later variants replaced each component in traditional pipelines with neural equivalents~\citep{arik2017deep, arik2017deep2,ren2019fastspeech,ren2020fastspeech,wang2019neural,ai2020neural}.
The first architectures which started using \dnns{}  exclusively as the vocoder were auto-regressive generative models, such as WaveNet~\citep{oord2016wavenet}, WaveRNN~\citep{kalchbrenner2018efficient}, SampleRNN~\citep{mehri2016samplernn}, Char2Wav~\citep{sotelo2017char2wav} or Tactron 2~\citep{shen2018natural}.

Due to their auto-regressive nature, these models do not leverage the parallel structure of modern hardware.
There have been several attempts to circumvent this problem:
One direction is to distill trained auto-regressive decoders into flow-based~\citep{kingma2016improving} convolutional student networks, as done by
Parallel WaveNet~\citep{oord2016wavenet} and Clarinet~\citep{ping2018clarinet}.
Another method is to utilize direct maximum likelihood training as done by several flow-based models, for example, WaveGlow~\citep{prenger2019waveglow}, FloWaveNet~\citep{kim2018flowavenet} or WaveFlow~\citep{ping2020waveflow}.
Other probabilistic approaches include those based on variational auto-encoders~\citep{oord2017neural,peng2020non} or diffusion probabilistic models~\citep{kong2020diffwave,chen2020wavegrad}.
Another family of methods is based on \gans{}~\citep{goodfellow2014generative}, examples include, \Mel{}~\citep{kumar2019melgan}, GAN-TTS~\citep{binkowski2019high}, WaveGAN~\citep{donahue2018adversarial}, HiFi-GAN~\citep{kong2020hifi}, Parallel WaveGAN~\citep{yamamoto2020pwg} or Multi-Band MelGAN~\citep{yang2021mbmelgan}. 

\subsection{Related Work}
\label{background:relatedwork}

Several previous proposals have collected Deepfake data:
The FaceForensics++ dataset~\citep{rossler2019faceforensics++} curated 1.8 million manipulated images and provides a benchmark for automated facial manipulation detection.
Celeb-DF~\citep{li2020celeb} contains high-quality face-swapping Deepfake videos of celebrities with more than 5,000 fake videos. 
\citet{dolhansky2020deepfake} released the Deepfake detection challenge that contains more than 100,000 videos, generated with different methods.

There exists a multitude of research into identifying GAN-generated images: Several approaches use CNNs in the image domain~\cite{marra2018detection,mo2018fake, yu2019attributing,tariq2019gan,wang2019cnn}, others use statistics in the image domain~\cite{mccloskey2018detecting,nataraj2019detecting}.
Another group of systems employs handcrafted features from the frequency domain: steganalysis-based features~\cite{marra2019gans}, spectral centroids~\cite{valle2018tequilagan} or frequency analysis~\citep{zhang2019detecting,durall2020watch,frank2020leveraging,qian2020thinking}. \citet{li2018exposing} proposed a CNN-based Deepfake video detection framework that utilizes artefacts that are consequences of the generation process. 
Another strain of research combines image analysis with audio analysis. 
\citet{chintha2020videodeepfake} combined a Deepfake detection with an audio spoofing detection to identify fake videos.
At the time of writing and to the best of our knowledge no work has provided an overview over Deepfake audio in isolation.

The signal processing community undertakes a related line of research.
The biyearly ASVspoof challenges~\citep{wu2017asvspoof, todisco2019asvspoof, nautsch19asvspoof} promote countermeasure against spoofing attacks that aim to fool speaker verification systems via different kinds of attacks.
Their benchmarking data sets include replay attacks, voice conversion, and synthesized audio files.
Note that the 2021 edition of the challenge features an audio Deepfake track but does not provide specific training data.
We imagine our data set to be used complementary with the training data of the challenge.
At the time of writing the 2021 edition is still on-going, but evaluating the best performing models in conjunction with our data set is an interesting direction for future work.
In the meantime, we adopt one of the baseline models of the ASVspoof challenge to enable a direct comparison.
This bi-yearly challenge has led to several proposed models for detecting spoofing attacks, for example, CNN-based methods~\citep{tom2018end,lavrentyeva2017audio,lai2019attentive}, ensemble methods on different feature representations~\citep{patil2017novel,kamble2018effectiveness,schonherr2017lol} or methods which detect unusual pauses in human speech~\citep{zhang2016voicelive,ahmed2020void}.
Additionally, another data set is proposed by Kinnunen~\etal~\citep{kinnunen2017reddots}.
They released a re-recorded version of the RedDots database for replay attack detection. \section{The data set}
\label{dataset}

In this Section we provide an overview of our data set.
It consists of 117,985 generated audio clips (16-bit PCM wav) and can be found on zenodo~\footnote{\, \href{https://zenodo.org/record/5642694}{\texttt{https://zenodo.org/record/5642694}} - DOI: 10.5281/zenodo.5642694}.
In total, it consists of approximately 196 hours of generated audio files.
We mostly base our work on the~\lj{}~\citep{ljspeech17} data set.
While~\tts{} models often get trained on private data sets, \lj{} is the most common public data set among the publication listed in Section~\ref{background:tts}.
Additionally, we consider the~\jsut{}~\citep{sonobe2017jsut} data set, a Japanese speech~corpus.

\paragraph{Reference data:}
We examine multiple networks trained on two reference data sets.
First, the \lj{}~\citep{ljspeech17} data set consisting of 13,100 short audio clips (on average 6 seconds each; roughly 24 hours total)
read by a female speaker.
It features passages from 7 non-fiction books, and the audio was recorded with a MacBook Pro microphone.
Second, we include samples based on the \jsut~\citep{sonobe2017jsut} data set, specifically, the basic5000 corpus.
This corpus consists of 5,000 sentences covering all basic kanji of the Japanese language.
(4.8 seconds on average; roughly 6.7 hours total).
A female native Japanese speaker performed the recordings in an anechoic room.
Note that we do not redistribute the reference data.
They are freely available online~\citep{ljspeech17,sonobe2017jsut}.

\clearpage
\paragraph{Architectures:}
We included a range of architectures in our data set:

\begin{itemize}[noitemsep,topsep=0pt,parsep=.5em,partopsep=0pt,leftmargin=2em]
    \item \textbf{MelGAN}: We include \Mel{}~\citep{kumar2019melgan}, which is one of the first GAN-based generative models for audio data.
    It uses a fully convolutional feed-forward network as the generator and operates on Mel spectrograms.
    The discriminator combines three different discriminators that operate on the original and two downsampled versions of the raw audio input.
    Additionally, it uses an auxiliary loss over the feature space of the three discriminators.
    \item \textbf{Parallel WaveGAN (\PWG{})}: WaveNet~\citep{oord2016wavenet} is one of the earliest and most common architectures,
    We include samples from one of its variants, Parallel WaveGAN~\citep{yamamoto2020pwg}.
    It uses the GAN training paradigm, with a non-autoregressive version of WaveNet as its generator.
    In a similar vein to \Mel{}, it uses an auxiliary loss, but in contrast, matches the \stft{} of the original training sample and the generated waveform over multiple resolutions.
\item \textbf{Multi-band MelGAN (\MBMel{})}: Incorporating more fine-grained frequency analysis, might lead to more convincing samples.
    We include \MBMel{}, which computes its auxiliary (frequency-based; inspired by~\PWG{}) loss in different sub-bands.
    Its generator is based on a bigger version of the \Mel{} generator. Still, instead of predicting the entire audio directly, the generator produces multiple sub-bands, which are then summed up to the complete audio signal.
    \item \textbf{Full-band MelGAN (\FBMel{})}: We include a variant of \MBMel{} which generates the complete audio directly and computes its auxiliary loss (the same as \PWG{}) over the full audio instead of its sub-bands.
    \item \textbf{HiFi-GAN (\hifi{})}: 
    \hifi{}~\citep{kong2020hifi} utilizes multiple sub-discriminators, each of which examines only a specific periodic part of the input waveform. 
    Similarly, its generator is built with multiple residual blocks, each observing patterns of different lengths in parallel.
    Additionally, \hifi{} employs the feature-space-based loss from \Mel{} and minimizes the $L_1$ distance between the Mel spectrogram of a generated waveform and a ground truth one in its loss function.
    \item \textbf{WaveGlow}: The training procedure might also influence the detectability of fake samples.
    Therefore, we include samples from \wav{} to investigate maximum-likelihood-based methods. 
    It is a flow-based generative model based on Glow~\citep{kingma2018glow}, whose architecture is heavily inspired by WaveNet.
\end{itemize}

Additionally, we examine \Mel{} both in a version similar to the original publication, which we denote as \Mel{}, and in a larger version with a bigger receptive field, \Mel{} (L)arge.
This version is similar to the one used by \FBMel{}, allowing for a one-to-one comparison.
Finally, we also obtain samples from a complete \tts{}-pipeline.
We use a conformer~\citep{gulati2020conformer} to map novel phrases (\ie, not part of \lj{}) to Mel spectrograms. Then we use a fine-tuned~\PWG{} model (trained on~\lj{}) to obtain the final audio.
We call this data set~\tts{}.
In total, we sample ten different data sets, seven based on \lj{} (\Mel{}, \Mel{} (L), \FBMel{}, \hifi{}, \wav{}, \PWG{}, \tts{}) and two based on \jsut{} (\MBMel{}, \PWG{}).

\begin{figure}[t!]
  \centering
  
  \begin{subfigure}{.49\columnwidth}
      \centering
      \includegraphics[width=.99\linewidth]{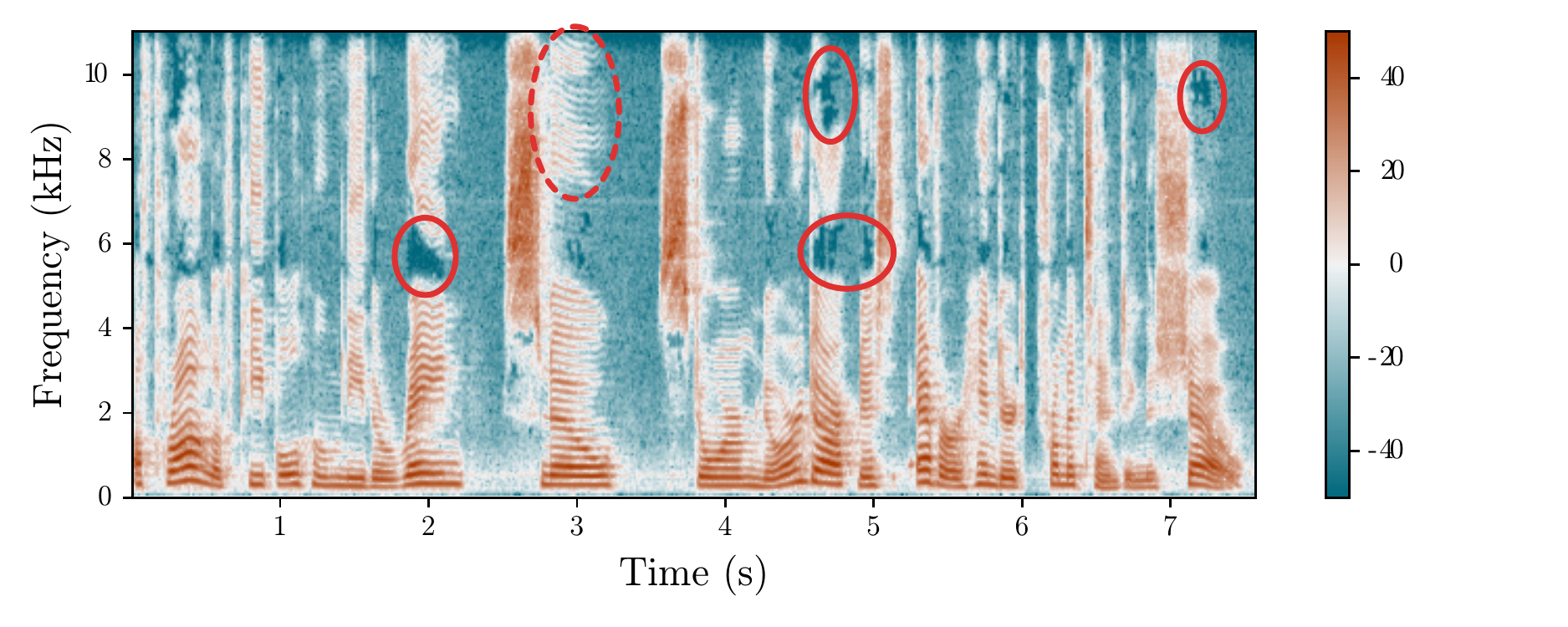}
      \caption{Original}
      \label{fig:specs_reduced:spec1}
  \end{subfigure}
\hfill
\begin{subfigure}{.49\columnwidth}
      \centering
      \includegraphics[width=.99\linewidth]{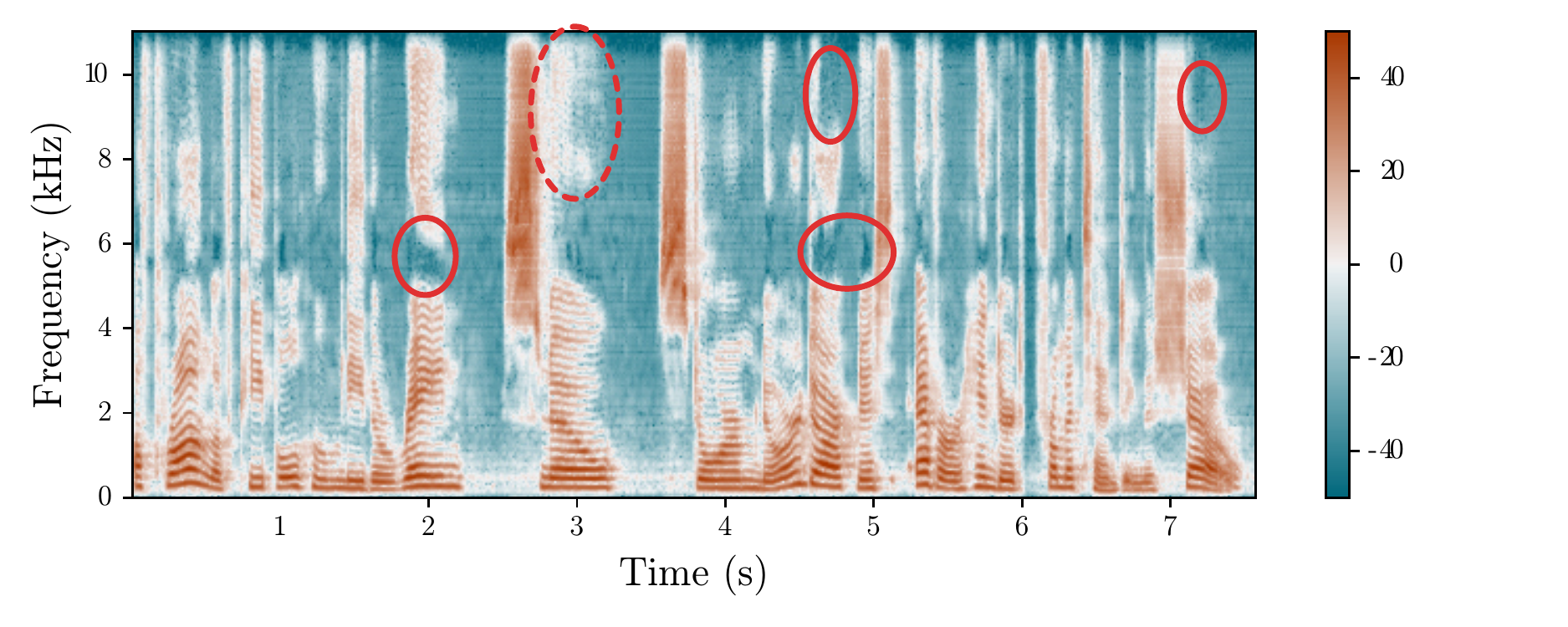}
      \caption{\Mel{}}
      \label{fig:specs_reduced:spec2}
  \end{subfigure}
  
\caption{\textbf{Spectrograms for the same sample, for different generating models.}
  They show the frequencies of a signal, plotted over the time of a signal.
  Lower frequencies at the bottom, higher at the top.
  Best viewed in color.
  }
  \vspace{-1em}
  \label{fig:specs_reduced}
\end{figure} \begin{figure}[t!]
  \centering
\begin{subfigure}{.99\columnwidth}
      \centering
      \includegraphics[width=.99\linewidth]{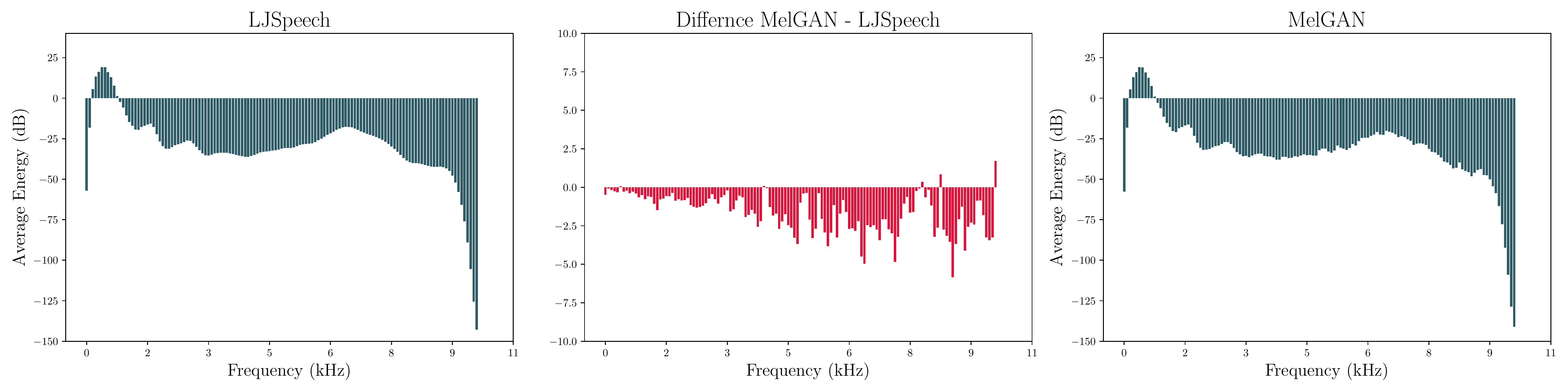}
      \label{fig:hist_reduced:melgan}
  \end{subfigure}
\caption{\textbf{Average energy per frequency bin.} 
  We show the average energy per frequency bin in dB.
  Additionally, we plot the difference to the original data (\lj{}).
  }
  \vspace{-1em}
  \label{fig:hist_reduced}
\end{figure} 

\paragraph{Sampling procedure:}
For \wav{} we utilize the official implementation~\citep{githubWaveglow} (commit 8afb643) in conjunction with the official pre-trained network on PyTorch Hub~\citep{githubTorchhub}.
HiFi-GAN also offers a public repository with pretrained models~\citep{githubHiFi}.
We use a popular implementation available on GitHub~\citep{githubPWG} (commit 12c677e) for the remaining networks.
When sampling the data set, we first extract Mel spectrograms from the original audio files, using the pre-processing scripts of the corresponding repositories.
We then feed these Mel spectrograms to the respective models to obtain the data set.
Intuitively, the networks are asked to ''recreate`` the original data sets.
For sampling the full~\tts{} results, we use the ESPnet project~\citep{watanabe2018espnet,hayashi2020espnet,inaguma-etal-2020-espnet,li2020espnet}.
To make sure the generated phrases do not overlap with the training set, we downloaded the common voices data set~\citep{ardila2019common} and extracted 16,283 phrases from it.

\begin{table}[t!]
    \centering
    \caption{
    \textbf{Basic statistics for all \lj{}-based data sets.}
    We report the average pitch frequency and its standard deviation as well as the average spectral centroid.
    }
    \resizebox{\linewidth}{!}{\begin{tabular}{@{}lrrrrrrrrr@{}}
        \toprule
\phantom{abc} & \multicolumn{1}{c}{\lj{}} & \multicolumn{1}{c}{\Mel{}} & \multicolumn{1}{c}{\Mel{} (L)} & \multicolumn{1}{c}{\FBMel{}} & \multicolumn{1}{c}{\MBMel{}} &\multicolumn{1}{c}{\hifi{}} &   \multicolumn{1}{c}{\wav{}} & \multicolumn{1}{c}{\PWG{}} \\ \midrule
\rule{0pt}{2.4ex}Avg. Pitch & $137.61$ & $133.51$ & $130.94$& $135.22$ & $133.519$ & $133.60$ & $135.80$ & $131.018$ \\
\rule{0pt}{2.4ex}Std. \phantom{c}Pitch & $49.64$ & $47.00$ & $46.15$ & $48.90$ & $48.19$ & $48.09$ & $47.38$ & $47.08$ \\
\rule{0pt}{2.4ex}Avg. Centroid & $2367.79$ & $2414.81$ & $2355.59$ & $2362.22$ & $2414.81$ & $2374.51$ & $2422.63$ & $2348.31$ \\
        \bottomrule
    \end{tabular}}
    \vspace{-1em}
    \label{tab:stats}
\end{table} \paragraph{Differences between the architectures:}
We perform an analysis of the differences between the architectures.
First, by plotting the spectrograms of an audio file in Figure~\ref{fig:specs_reduced} (\lj{} \datasetstyling{008-0217}; all data sets can be found in Section~6 of the supplementary material).
Generally, all architectures produce spectrograms different from the original.
The networks seem to struggle with the absence of information generally (solid circles in Figure~\ref{fig:specs_reduced:spec1}) as well as with higher frequencies, especially for vocals (dashed circle).
Additionally, \Mel{} and \wav{} seem to cause a repeating horizontal pattern.
The remaining networks (all using an auxiliary loss in the frequency domain) do not seem to exhibit this behavior.
However, they still produce apparent differences.
Note that these differences are visible when plotting the audio but generally inaudible when listening to the samples.

Second, we perform a prosody analysis of each data set.
We collect 10.000 samples from \lj{} and the corresponding sample from each of our architectures.
For each data set, we compute the fundamental frequency (pitch) by using normalized cross-correlation and median smoothing~\citep{huang-79-fast} in the range $50-500$Hz.
Additionally, we compute the center of mass of the frequency spectrum by the mean of the frequencies weighted by their magnitudes (the so-called spectral centroid).
The results can be found in Table~\ref{tab:stats} and confirm our visual observations.
While all architectures come close to the original, none can approximate it perfectly.
Generally, all vocoders produce a lower and less varied pitch. 

The spectral centroid results are varied.
To investigate further, we perform an additional, more fine-grained analysis by plotting a histogram of the energy contained in each frequency bin.
Furthermore, we plot the relative difference to the original data, i.e., the difference weighted by the initial (\lj{}) energy.
The plots can be found in Figure~\ref{fig:hist_reduced}.
For brevity, we only show the \Mel{} comparison here, the other histograms can be found in Section~6 of the supplementary material. 
However, all analyses had similar results. The histograms' overall shape is identical, but the generated samples exhibit apparent differences, especially in the higher frequencies.

\paragraph{Licensing:}
The \lj data set is in the public domain.
The \jsut corpus is licensed by CC-BY-SA 4.0, noting that redistribution is only permitted in some instances.
We contacted the author, who saw no conflict in distributing our fake samples, as long as it's for research purposes.
Thus, we also released our data set under a CC-BY-SA 4.0 license.

\paragraph{Ethical considerations:}
Our data set consists of phrases from non-fiction books (\lj{}) and everyday conversational Japanese (\jsut{}), which are already available online.
The same is true for all models used to generate this data set.
One might wonder if releasing research into detecting Deepfakes might negatively affect the detection ''arms race``.
This discussion has a long-standing history in the security community, and the general conclusion is that withholding research is hurtful.
We provide a more in-depth discussion of this topic from the perspective of security researchers in the supplementary material (Section~1).

 \section{Providing a baseline}

We base our experiment on the ASVspoofchallenge~\citep{todisco2019asvspoof} introduced in Section~\ref{background:relatedwork}.
 The challenge aims to promote research into detecting (audio) spoofing attacks and speaker verification.
As a point of reference, the challenge offers two baseline models (\ac{GMM} and \rawnet~\citep{jung2020improved,tak2021rawnet2}).
We adopt these models and the metric used by the challenge to compare the two domains.
 
\subsection{Experiments}
\label{eval:results}

We start by training seven different classifiers, one for each vocoder in our data set (\Mel{}, \Mel{} (L), \FBMel{}, \MBMel{}, \hifi{}, \PWG{} and \wav{}).
For training our classifiers, we exclusively use the data sets based on~\lj{}.
We use the~\jsut{} (different speaker, language, and recording setup) and~\tts{} (same speaker, completely novel phrases) data sets for accessing the classifiers ability to generalize to an unknown setting.
We train six additional models in a leave-one-out setting to control if the models picked up on vocoder-specific characteristics. 
Finally, we simulate a phone recording to emulate a real-world fraud attempt.

For each classifier, we evaluate the performance on all vocoders over a hold-out set of 20\% of the data.
We use the \ac{EER} as our evaluation metric.
The ASVspoof challenge also uses this metric. 
It is defined as the point on the \acs{ROC} curve, where false acceptance rate and false rejection rate are equal and is commonly used to assess the performance of binary classifications tasks like biometric security systems~\cite{scheuermann2000usability}.
The best possible value is 0.0 (no wrong predictions), worst 1.0 (everything wrong), guessing is 0.5. 
Additionally, we compute \emph{average} \acl{EER} (aEER) over all test sets.

When training \gmm{} models, we follow~\citet{sahidullah2015comparison} and train two \gmms{} per data set, one fitting the real distribution (the original~\lj{} data set) and one fitting the generated audio samples (the respective vocoder-samples from our data set).
In addition to the \ac{LFCC} features used by~\citet{sahidullah2015comparison}, we evaluate \ac{MFCC} features. To classify a given sample, we calculate its likelihood~$\Lambda(\mathbf{X})$~via
\begin{equation}
    \Lambda(\mathbf{X}) = \log p(\mathbf{X}|\theta_n) - \log p(\mathbf{X}|\theta_s),
\end{equation}
where $\mathbf{X}$ are the input features (namely \mfcc{} or \lfcc{}) and $\theta_n$ and $\theta_s$ are the \ac{GMM} model parameter of the real and the generated audio distributions, respectively.
The out-of-distribution models are exclusively trained on~\lfcc{} features, since we found them to strictly outperform the \mfcc{} features (\cf, Section~\ref{tab:lfcc_results}).

Additionally, we train \rawnet~\citep{jung2020improved} instances to investigate a neural alternative.
\rawnet{} is a CNN-GRU hybrid model which extracts a speaker embedding directly from raw audio.
When used to perform speaker verification (or Deepfake detection), a fully connected layer is trained on top of this embedding to make the final decision~\citep{tak2021rawnet2}.
Details on all training setups can be found in the supplementary material (Section~3).

\paragraph{Single training set:}
In a first experiment, we evaluate the performance when training on a single data set.
For the \gmm{} experiments, we only present the \lfcc{} results since we found them to outperform the \mfcc{} features strictly.
\lfcc{} features contain a significantly higher amount of high-frequency components. We hypothesize that these are meaningful for achieving good overall performance.
Similar patterns were observed in the image domain~\citep{frank2020leveraging}, implying that methods might transfer between the two.
The results are presented in Table~\ref{tab:lfcc_results}.
The rows show the respective training sets and the columns the different test sets.
Gray values indicate that the same generative model is used for the training of the \ac{GMM} classifier as for the test set.

When training on a single data set, we observe that \FBMel{} serves as the best prior for all other data sets, achieving the lowest average \ac{EER} ($0.062$).
Intuitively this makes sense.
\FBMel{} uses the same architecture as \Mel{} (L), while deploying a similar auxiliary loss as \PWG{} or \MBMel{}.
Generally, we can see a clear divide between \Mel{}/\Mel{} (L) and the other networks, which we will explore in Section~\ref{eval:attribution}.

\begin{table}[t!]
    \centering
    \caption{
    \textbf{Equal Error Rate (EER) of the baseline classifier on different subset (LFCC).}
    We train a new GMM model for each data set and compute the EER as well as the \textbf{aEER}.
    }
    \resizebox{\linewidth}{!}{\begin{tabular}{@{}lrrrrrrrrrrrr@{}}
        \toprule
 &  \multicolumn{8}{c}{\lj} & \phantom{a} &\multicolumn{2}{c}{\jsut}\\ 
\cmidrule{2-9} \cmidrule{11-12} \rule{0pt}{2.2ex}
Training Set & \multicolumn{1}{c}{\Mel{}} & \multicolumn{1}{c}{\Mel{} (L)} & \multicolumn{1}{c}{\FBMel{}} & \multicolumn{1}{c}{\MBMel{}} &   \multicolumn{1}{c}{\hifi{}} &\multicolumn{1}{c}{\wav{}} & \multicolumn{1}{c}{\PWG{}} & \multicolumn{1}{c}{\FTTS{}} & & \multicolumn{1}{c}{\MBMel{}} & \multicolumn{1}{c}{\PWG{}} & \multicolumn{1}{c}{\phantom{a}\textbf{aEER}} \\ \midrule 
\rule{0pt}{2.4ex}\Mel{} & \gray{0.148} & \textbf{0.094} & 0.155 & 0.153 & 0.168 & 0.189 & 0.109 & 0.023 &  & 0.384 & 0.533 & 0.215 \\
\rule{0pt}{2.4ex}\Mel{} (L) & \textbf{0.119} & \gray{0.044} & 0.176 & 0.132 & 0.150 & 0.245 & 0.115 & \textbf{0.012} &  & 0.406 & 0.607 & 0.222 \\
\rule{0pt}{2.4ex}\MBMel{} & 0.359 & 0.316 & \gray{0.002} & 0.124 & 0.083 & 0.007 & \textbf{0.017} & 0.021 &  & 0.017 & 0.051 & 0.108 \\
\rule{0pt}{2.4ex}\FBMel{} & 0.197 & 0.133 & 0.030 & \gray{0.025} & \textbf{0.034} & 0.037 & 0.019 & 0.025 &  & 0.026 & 0.058 & \textbf{0.062} \\
\rule{0pt}{2.4ex}\hifi{} & 0.255 & 0.193 & 0.034 & \textbf{0.050} & \gray{0.029} & 0.035 & 0.020 & 0.018 &  & 0.057 & 0.123 & 0.089 \\
\rule{0pt}{2.4ex}\PWG{} & 0.402 & 0.374 & 0.008 & 0.161 & 0.100 & \gray{0.001} & \textbf{0.017} & 0.042 &  & \textbf{0.014} & \textbf{0.042} & 0.124 \\
\rule{0pt}{2.4ex}\wav{} & 0.287 & 0.237 & \textbf{0.015} & 0.066 & 0.041 & \textbf{0.008} & \gray{0.003} & 0.015 &  & 0.031 & 0.075 & 0.085 \\

        \bottomrule
    \end{tabular}}
    \begin{flushleft}
    \scriptsize{When the distribution is part of the training set we highlight it in \gray{gray}. For other results, we highlight the best results per column in \textbf{bold}.}
    \end{flushleft}
    \label{tab:lfcc_results}
    \vspace{-1.5em}
\end{table}
 
\begin{table}[t!]
    \centering
    \caption{
    \textbf{Equal Error Rate (EER) of the \rawnet{} classifier.}
    We train a single \rawnet{} model per data set and compute the EER as well as the \textbf{aEER}.
    }
    \resizebox{\linewidth}{!}{\begin{tabular}{@{}lrrrrrrrrrrrr@{}}
        \toprule
 &  \multicolumn{8}{c}{\lj} & \phantom{a} &\multicolumn{2}{c}{\jsut}\\ 
\cmidrule{2-9} \cmidrule{11-12} \rule{0pt}{2.2ex}
Training Set & \multicolumn{1}{c}{\Mel{}} & \multicolumn{1}{c}{\Mel{} (L)} & \multicolumn{1}{c}{\FBMel{}} & \multicolumn{1}{c}{\MBMel{}} &   \multicolumn{1}{c}{\hifi{}} &\multicolumn{1}{c}{\wav{}} & \multicolumn{1}{c}{\PWG{}} & \multicolumn{1}{c}{\FTTS{}} & & \multicolumn{1}{c}{\MBMel{}} & \multicolumn{1}{c}{\PWG{}} & \multicolumn{1}{c}{\phantom{a}\textbf{aEER}} \\ \midrule 
\rule{0pt}{2.4ex}\Mel{} & \gray{0.001} & \textbf{0.001} & 0.485 & 0.509 & 0.525 & 0.497 & 0.407 & 0.356 &  & 0.113 & 0.089 & 0.292 \\
\rule{0pt}{2.4ex}\Mel{} (L) & \textbf{0.008} & \gray{0.000} & 0.511 & 0.490 & 0.486 & 0.369 & 0.446 & 0.265 &  & \textbf{0.009} & \textbf{0.003} & \textbf{0.258} \\
\rule{0pt}{2.4ex}\MBMel{} & 0.118 & 0.371 & \gray{0.003} & 0.282 & \textbf{0.216} & 0.302 & \textbf{0.002} & 0.522 &  & 0.922 & 0.997 & 0.357 \\
\rule{0pt}{2.4ex}\FBMel{} & 0.161 & 0.239 & 0.122 & \gray{0.082} & 0.304 & \textbf{0.259} & 0.130 & 0.391 &  & 0.974 & 0.994 & 0.363 \\
\rule{0pt}{2.4ex}\hifi{} & 0.174 & 0.437 & 0.242 & 0.364 & \gray{0.023} & 0.359 & 0.057 & 0.098 &  & 0.499 & 0.719 & 0.319 \\
\rule{0pt}{2.4ex}\PWG{} & 0.052 & 0.358 & 0.261 & \textbf{0.234} & 0.324 & \gray{0.000} & 0.006 & \textbf{0.001} &  & 0.984 & 0.999 & 0.358 \\
\rule{0pt}{2.4ex}\wav{} & 0.086 & 0.379 & \textbf{0.07}9 & 0.361 & 0.226 & 0.316 & \gray{0.001} & 0.250 &  & 0.409 & 0.786 & 0.294 \\
        \bottomrule
    \end{tabular}}
    \begin{flushleft}
    \scriptsize{When the distribution is part of the training set we highlight it in \gray{gray}. For other results, we highlight the best results per column in \textbf{bold}.}
    \end{flushleft}
    \label{tab:rawnet_single}
    \vspace{-1.5em}
\end{table} 
When examining completely novel data (\jsut{}), all classifier drop in performance.
However, \PWG{}, \wav{}, \hifi{}, \FBMel{} and, \MBMel{} still serve as a good prior, implying that the generating architectures exhibit common patterns which can be recognized for different training data sets and speakers.
Again, a similar pattern was also observed in the image domain~\citep{wang2019cnn}.
The~\tts{} data set is one of the easiest data sets.
This undermines our belief that data that recreates the training set is harder to classify correctly.
Interestingly the \PWG{} classifier does not generalize well to the~\tts{} data set.
Remember that while we use completely novel phrases, the vocoder for this data set is a \PWG{} network trained on~\lj{}.
This might imply that our models overfit their specific training set.

This trend can also be seen in the \rawnet{} results, which overall perform worse than the \gmm{} models.
They seem to overfit their respective training architecture, which prevents them from generalizing to other data sets.
This explains the good performance of the \Mel{}/\Mel{} (L) models and the \PWG{}/\tts{} models, since these pairs share generator architectures.
Additionally, we can note that the \Mel{}/\Mel{} (L) models serve as a good prior for generalizing to the \jsut{} data sets.

\paragraph{Leave-one-out:}
We explore this hypothesis by running a leave-one-out experiment. Results can be found in Table~\ref{tab:leave_one_out_results}.
Overall the results improve on the aEER ($0.062 \to 0.058$).
Also, the generalization results to a novel setting (\jsut{}) increase significantly.
However, \FBMel{} seems to be an essential ingredient for good performance on the \jsut{} data since not training on it hurts performance significantly.
Additionally, the~\Mel{} and~\Mel{} (L) data sets still prove to be a challenge, even when included in the training set.

\begin{table}[t!]
    \centering
    \caption{
        \textbf{Equal Error Rate (EER) for the \gmm{} classifier in an out-of-distribution setting.}
        We train a new GMM model for each but one distribution on~\lfcc{} features.
    }
    \resizebox{\linewidth}{!}{\begin{tabular}{@{}lrrrrrrrrrrrr@{}}
        \toprule
 &  \multicolumn{8}{c}{\lj} & \phantom{a} &\multicolumn{2}{c}{\jsut}\\ 
\cmidrule{2-9} \cmidrule{11-12} \rule{0pt}{2.2ex}
Training Set & \multicolumn{1}{c}{\Mel{}} & \multicolumn{1}{c}{\Mel{} (L)} & \multicolumn{1}{c}{\FBMel{}} & \multicolumn{1}{c}{\MBMel{}} &   \multicolumn{1}{c}{\hifi{}} &\multicolumn{1}{c}{\wav{}} & \multicolumn{1}{c}{\PWG{}} & \multicolumn{1}{c}{\FTTS{}} & & \multicolumn{1}{c}{\MBMel{}} & \multicolumn{1}{c}{\PWG{}} & \multicolumn{1}{c}{\phantom{a}\textbf{aEER}} \\ \midrule 
\rule{0pt}{2.4ex}\Mel{} & 0.220 & \gray{0.146} & \gray{0.009} & \gray{0.051} & \gray{0.040} & \gray{0.016} & \gray{0.009} & 0.006 &  & 0.023 & 0.067 & 0.065 \\
\rule{0pt}{2.4ex}\Mel{} (L) & \gray{0.231} & 0.164 & \gray{0.010} & \gray{0.045} & \gray{0.040} & \gray{0.014} & \gray{0.012} & 0.009 &  & 0.013 & \textbf{0.043} & 0.064 \\
\rule{0pt}{2.4ex}\MBMel{} & \gray{0.187} & \gray{0.117} & 0.013 & \gray{0.043} & \gray{0.039} & \gray{0.018} & \gray{0.010} & \textbf{0.002} &  & 0.058 & 0.141 & 0.069 \\
\rule{0pt}{2.4ex}\FBMel{} & \gray{0.191} & \gray{0.116} & \gray{0.013} & 0.058 & \gray{0.053} & \gray{0.022} & \gray{0.013} & 0.003 &  & 0.084 & 0.220 & 0.086 \\
\rule{0pt}{2.4ex}\hifi{} & \gray{0.192} & \gray{0.119} & \gray{0.011} & \gray{0.050} & 0.047 & \gray{0.015} & \gray{0.012} & 0.004 &  & \textbf{0.020} & 0.061 & \textbf{0.058} \\
\rule{0pt}{2.4ex}\PWG{} & \gray{0.176} & \gray{0.105} & \gray{0.014} & \gray{0.044} & \gray{0.042} & 0.034 & \gray{0.013} & 0.005 &  & 0.033 & 0.101 & 0.062 \\
\rule{0pt}{2.4ex}\wav{} & \gray{0.191} & \gray{0.114} & \gray{0.013} & \gray{0.049} & \gray{0.045} & \gray{0.021} & 0.015 & 0.008 &  & 0.031 & 0.078 & 0.062 \\

        \bottomrule
    \end{tabular}}
\begin{flushleft}
    \scriptsize{When the distribution is part of the training set we highlight it in \gray{gray}. For other results, we highlight the best results per column in \textbf{bold}.}
    \end{flushleft}
    \label{tab:leave_one_out_results}
    \vspace{-1.5em}
\end{table}

\begin{table}[t!]
    \centering
    \caption{
    \textbf{Equal Error Rate (EER) of the \rawnet{} classifier in an out-of-distribution setting.}
    We train a single \rawnet{} model on all but one distribution and compute the EER as well as the \textbf{aEER}.
    }
    \resizebox{\linewidth}{!}{\begin{tabular}{@{}lrrrrrrrrrrrr@{}}
        \toprule
 &  \multicolumn{8}{c}{\lj} & \phantom{a} &\multicolumn{2}{c}{\jsut}\\ 
\cmidrule{2-9} \cmidrule{11-12} \rule{0pt}{2.2ex}
Training Set & \multicolumn{1}{c}{\Mel{}} & \multicolumn{1}{c}{\Mel{} (L)} & \multicolumn{1}{c}{\FBMel{}} & \multicolumn{1}{c}{\MBMel{}} &   \multicolumn{1}{c}{\hifi{}} &\multicolumn{1}{c}{\wav{}} & \multicolumn{1}{c}{\PWG{}} & \multicolumn{1}{c}{\FTTS{}} & & \multicolumn{1}{c}{\MBMel{}} & \multicolumn{1}{c}{\PWG{}} & \multicolumn{1}{c}{\phantom{a}\textbf{aEER}} \\ \midrule 
\rule{0pt}{2.4ex}\Mel{} & 0.008 & \gray{0.005} & \gray{0.023} & \gray{0.137} & \gray{0.098} & \gray{0.076} & \gray{0.011} & 0.019 &  & \textbf{0.000} & \textbf{0.000} & \textbf{0.040} \\
\rule{0pt}{2.4ex}\Mel{} (L) & \gray{0.005} & 0.046 & \gray{0.009} & \gray{0.048} & \gray{0.050} & \gray{0.024} & \gray{0.004} & 0.020 &  & 0.985 & 0.996 & 0.241 \\
\rule{0pt}{2.4ex}\MBMel{} & \gray{0.013} & \gray{0.039} & 0.037 & \gray{0.102} & \gray{0.060} & \gray{0.055} & \gray{0.005} & 0.089 &  & 0.860 & 0.758 & 0.214 \\
\rule{0pt}{2.4ex}\FBMel{} & \gray{0.013} & \gray{0.023} & \gray{0.032} & 0.216 & \gray{0.058} & \gray{0.054} & \gray{0.011} & 0.026 &  & 0.092 & 0.088 & 0.065 \\
\rule{0pt}{2.4ex}\hifi{} & \gray{0.006} & \gray{0.009} & \gray{0.031} & \gray{0.113} & 0.196 & \gray{0.065} & \gray{0.010} & 0.044 &  & 0.001 & 0.001 & 0.048 \\
\rule{0pt}{2.4ex}\PWG{} & \gray{0.005} & \gray{0.004} & \gray{0.026} & \gray{0.108} & \gray{0.088} & 0.209 & \gray{0.011} & 0.044 &  & 0.047 & 0.123 & 0.069 \\
\rule{0pt}{2.4ex}\wav{} & \gray{0.001} & \gray{0.006} & \gray{0.008} & \gray{0.038} & \gray{0.046} & \gray{0.010} & 0.001 & \textbf{0.005} &  & 0.828 & 0.904 & 0.205 \\
        \bottomrule
    \end{tabular}}
    \begin{flushleft}
    \scriptsize{When the distribution is part of the training set we highlight it in \gray{gray}. For other results, we highlight the best results per column in \textbf{bold}.}
    \end{flushleft}
    \label{tab:ood_raw_net}
    \vspace{-1.5em}
\end{table} 
The results are similar for \rawnet{} (Table~\ref{tab:ood_raw_net}).
When trained on multiple distributions, the networks can successfully generalize, even surpassing the best aEER ($0.04$).
However, some models still overfit to the training data, making generalization to \jsut{} impossible (\Mel{} (L), \MBMel{}, \wav{}).
Additionally, the better average performance is traded off with worse performance on the training data.
For example, the best performing average model has a 13\% false acceptance/false rejection rate.
This would be unacceptable in a real-life setting.

\paragraph{Simulating a phone call:}
Finally, we return to our motivating example and examine how well our models generalize to a (simulated) real-life setting.
We emulate a phone recording for the three test data set (both \jsut{} data sets and the full \tts{}-pipeline) and evaluate the out-of-distribution models on them.
The \gmm{} classifiers work exceptionally well, even surpassing the performance in the out-of-distribution setting.
The highest \ac{EER} we could detect was $0.003$, all other results were lower or separated the data perfectly (a full table can be found in Section~5 of the supplementary material).
The results are flipped for the \rawnet{} models.
While in the clean setting, the (some) models classify the data almost perfectly, under the phone simulation, the error rate shoots up significantly.

While we only examined a simulated setting, we take this as the first evidence that our data set can be used to extrapolate classifier performance to the real world.
Overall, we can conclude that these first results are encouraging, but there is still much room for improvement.

\subsection{Attribution}
\label{eval:attribution}

\begin{figure}[t!]
  \centering
\begin{subfigure}{.49\columnwidth}
      \centering
      \includegraphics[width=.99\linewidth]{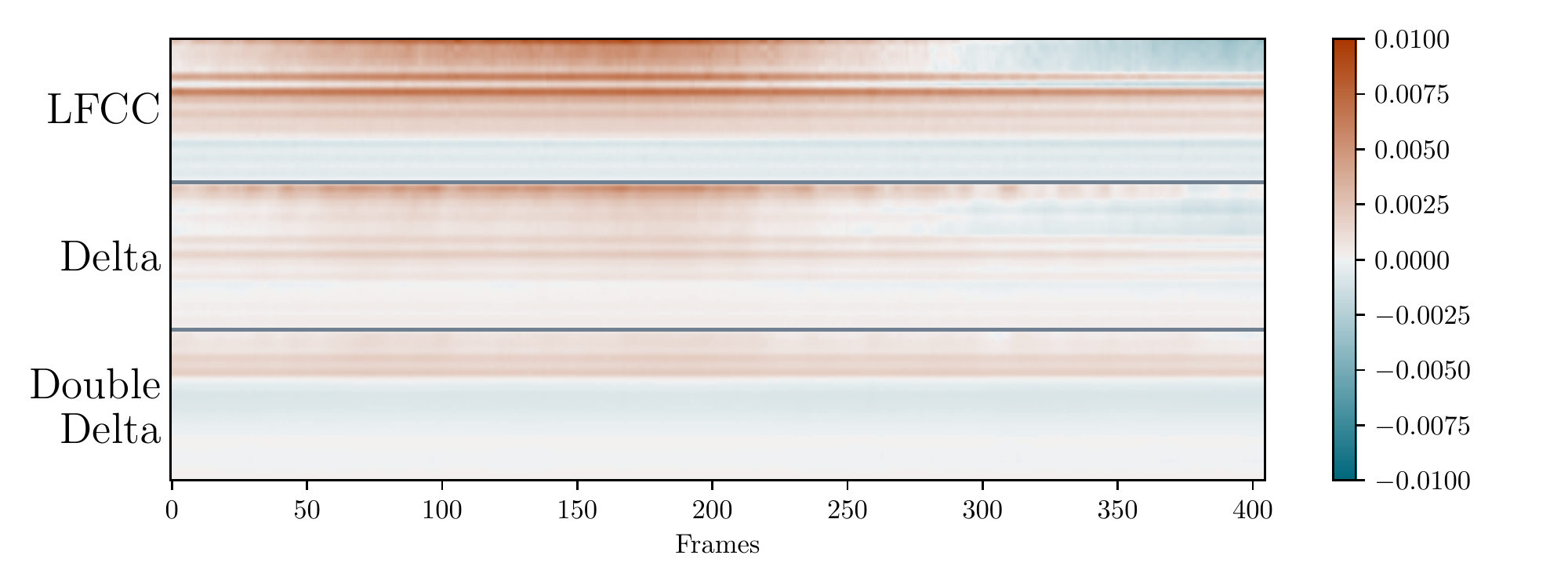}
      \caption{\Mel{} (L)}
      \label{fig:attribution1}
  \end{subfigure}
\begin{subfigure}{.49\columnwidth}
      \centering
      \includegraphics[width=.99\linewidth]{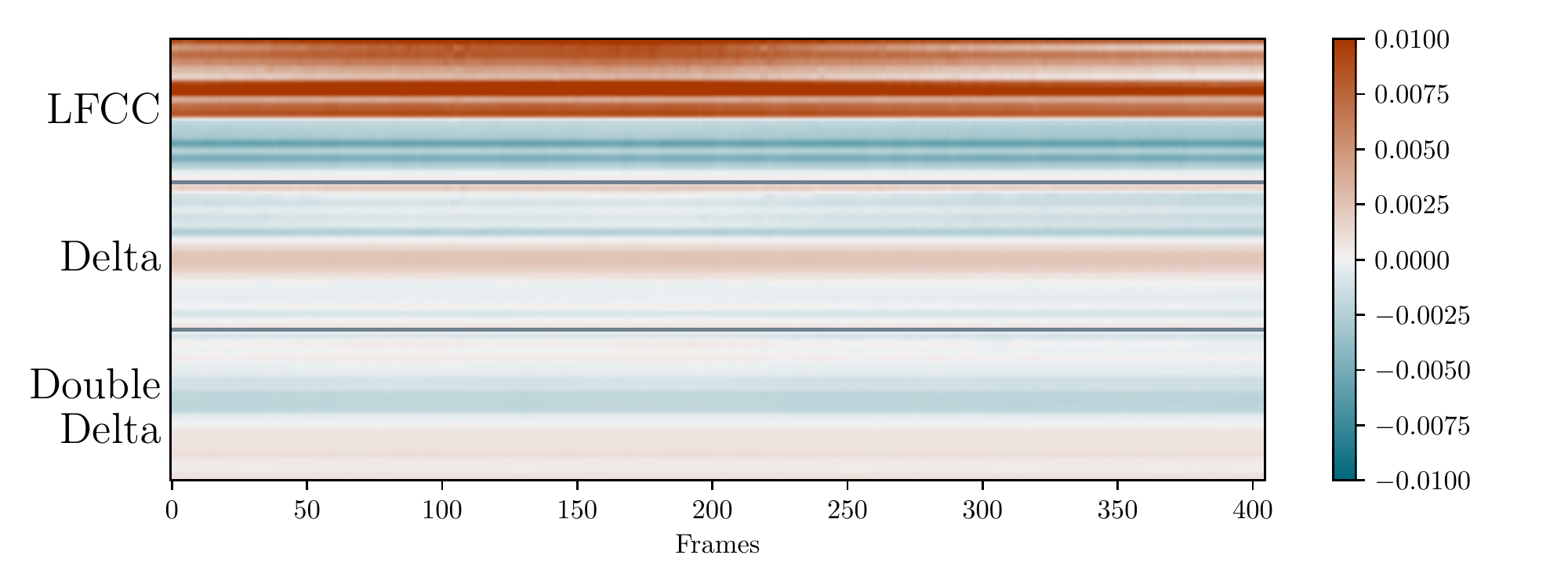}
      \caption{\FBMel{}}
      \label{fig:attribution2}
  \end{subfigure}
\hfill
\begin{subfigure}{.49\columnwidth}
      \centering
      \includegraphics[width=.99\linewidth]{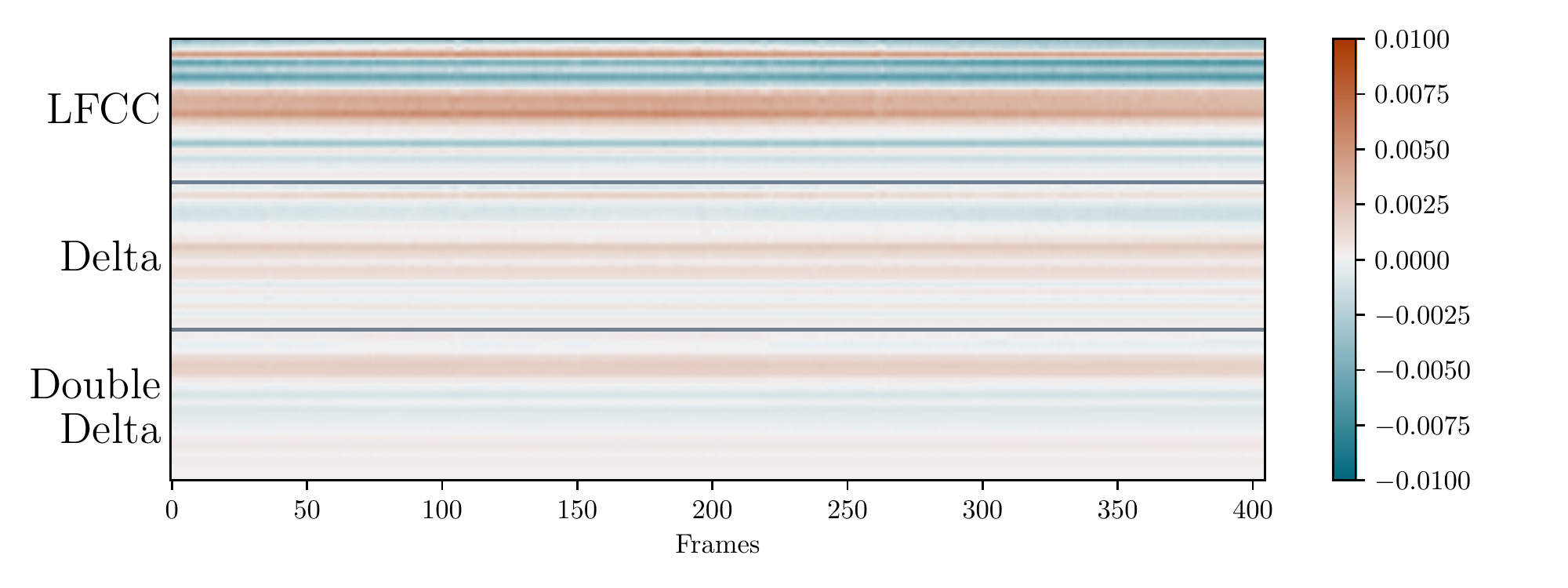}
      \caption{\MBMel{}}
      \label{fig:attribution3}
  \end{subfigure}
\begin{subfigure}{.49\columnwidth}
      \centering
      \includegraphics[width=.99\linewidth]{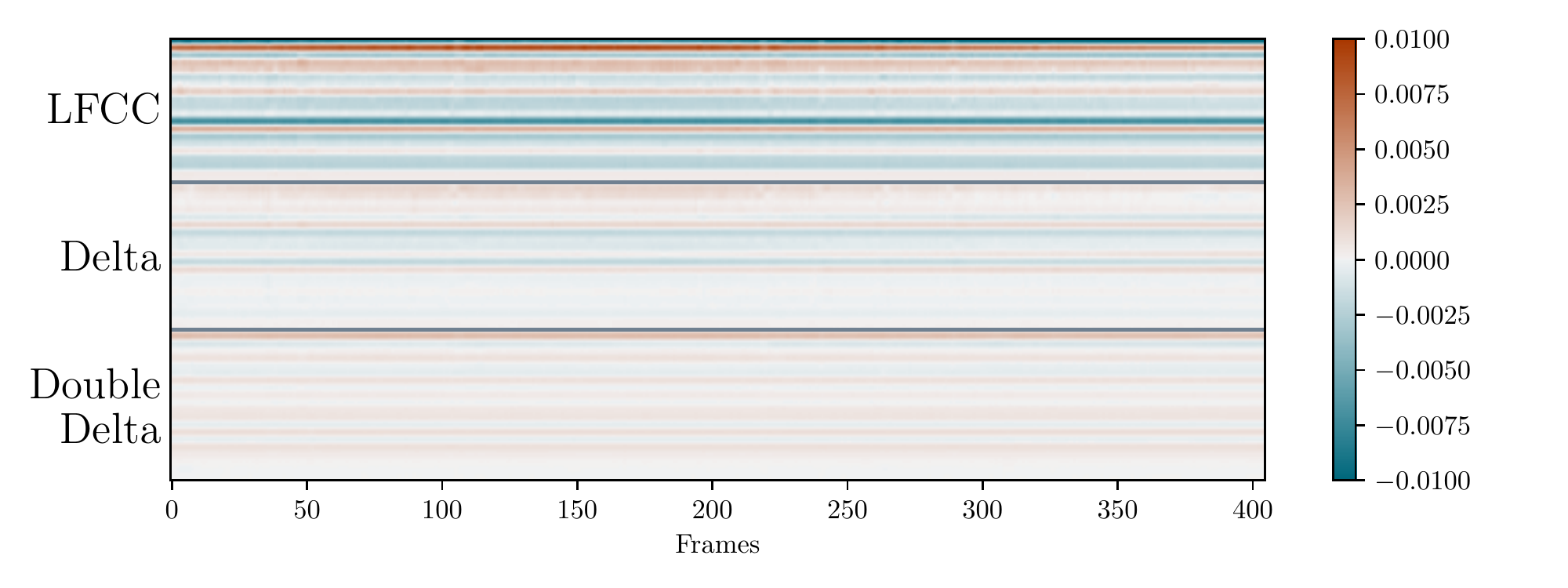}
      \caption{\PWG{}}
      \label{fig:attribution4}
  \end{subfigure}
\caption{\textbf{Attribution of the different models on a real audio sample.} 
  We show the \lfcc{}, delta, and, double delta features.
The plot can be read similarly to the spectrogram plots, \ie, features computed over lower frequencies are at the bottom of their respective section, features over higher frequencies are at the top.
  Best viewed in color.
  }
  \label{fig:attribution}
  \vspace{-1em}
\end{figure}

Finally, we want to investigate which parts of the audio signal influence the prediction.
To this end, we implemented BlurIG~\citep{xu2020attribution}, a popular attribution method.
We inspect the attribution of four classifiers for the audio clip used in Section~\ref{dataset}.
The results are displayed in Figure~\ref{fig:attribution}.

Overall, we can see a shift from very broad attention, spread somewhat evenly across all three feature representations (\Mel{} (L)), to more narrow-focused attention across very specific filters (\PWG{}).
\Mel{} (L) and \FBMel{} classifiers operate (mostly) on the higher frequencies, while \MBMel{} and \PWG{} also rely on low frequencies for the detection.
These observations confirm our suspicion about the \mfcc{} features.
They mask higher frequencies, needed for classifying \Mel{} (L) and \FBMel{}, while over-representing lower frequencies, which still leads to a good performance on the \MBMel{} and \PWG{} data sets.
This also explains the performance of \FBMel{}, which strikes a balance between all necessary features.
The spread-out attribution might also explain the poor in-distribution performance of the classifiers trained on the \Mel{} variants since the classifier needs to focus on a broader range of features.

 \section{Discussion}
\label{discussion}

In this paper, we took the first step towards research into audio Deepfakes.
While we hope our data set proves useful for future practitioners, there are several limitations to our work: 

\paragraph{Evaluating on realistic data:}
The difficulties of obtaining realistic data set have been a long-standing problem in the security community~\citep{sommer2010outside}.
Often benign data is readily available, but data used in malicious contexts is hard to come by.
That leaves us with estimating real-world performance on proxy data.
We argue that in our case, we might have good odds that results transfer.
Currently, images generated by off-the-shelf neural networks are used in malicious attempts~\citep{burgess2020telegram}.
We expect the number of audio Deepfakes to increase as well.

\paragraph{Variety of the data:}
We specifically choose to focus on the~\lj{} corpus since it is commonly used for training generative audio models.
That allows a one-to-one comparison.
However, it only contains recordings by one speaker.
We can make some observations about generalization by comparing against the~\jsut{} and~\tts{} data sets, but a broader analysis focusing on different scenarios would be ideal.

\paragraph{Adversarial examples and perturbations:}
Deepfake-image detectors have already been shown to be vulnerable against adversarial examples~\citep{carlini2020evading}.
There also exists a myriad of adversarial attacks against automatic speech recognition~\citep{carlini-18-audio,schoenherr-19-psychoacoustics,yuan-18-commandersong,schoenherr-19-psychoacoustics,alzantot-18-did,schoenherr-20-imperio,aghakhani-20-venomave} (\citet{abdullah-20-sok} provide a survey).
We have looked at phone recordings, but classifiers should report their robustness against these attacks and other common perturbations (noise, room responses, over-the-air settings, etc.) as part of their evaluation.
In this work, we focused on providing the first steps towards audio Deepfake detection.

\section{Conclusion}
\label{conclusion}

This paper presents a starting point for researchers who want to investigate generated audio signals.
We started by presenting a broad overview of signal processing techniques and common feature representations as well as a survey of the current \tts{} landscape.
We then moved on to introduce our main contribution, a novel data set, with samples from six different state-of-the-art architectures across two languages.
We discovered subtle differences between the different models by plotting the frequency spectrum, especially among the higher frequencies.
Following up, we conducted a prosody analysis and investigated each data set's average energy per frequency. 
This analysis confirmed our previous findings, revealing that while all models come close to correctly approximating the training data, we can still detect differences unique to each model.
To provide a baseline for future practitioners, we trained several baseline models. We evaluated their performance across the different data sets and multiple settings.
Specifically, we trained \gmm{} and neural network-based solutions.
While we found the neural networks to perform better overall, the \gmm{} classifiers proved to be more robust, which might give them an advantage in real-life settings.
Finally, we inspected the different classifiers using an attribution method. We found that lower frequencies cannot be neglected while high-frequency information proved indispensable. \clearpage

\begin{ack}
We would like to thank our colleagues Thorsten Eisenhofer, Thorsten Holz, Dorothea Kolossa, and our anonymous reviewers for their valuable feedback and fruitful discussions. 
Additionally, we would like to thank Tomoki Hayashi, Hemlata Tak, and the \wav{} team for their excellent repositories.
This work was supported by the Deutsche Forschungsgemeinschaft (DFG, German Research Foundation) under Germany's Excellence Strategy -- EXC-2092  \textsc{CaSa} -- 390781972.
\end{ack}

\bibliographystyle{plainnat}
\bibliography{output.bbl}

\begin{thebibliography}{117}
\providecommand{\natexlab}[1]{#1}
\providecommand{\url}[1]{\texttt{#1}}
\expandafter\ifx\csname urlstyle\endcsname\relax
  \providecommand{\doi}[1]{doi: #1}\else
  \providecommand{\doi}{doi: \begingroup \urlstyle{rm}\Url}\fi

\bibitem[Abdullah et~al.(2020)Abdullah, Warren, Bindschaedler, Papernot, and
  Traynor]{abdullah-20-sok}
Hadi Abdullah, Kevin Warren, Vincent Bindschaedler, Nicolas Papernot, and
  Patrick Traynor.
\newblock {SoK: The Faults in our ASRs: An Overview of Attacks against
  Automatic Speech Recognition and Speaker Identification Systems}.
\newblock In \emph{IEEE Symposium on Security and Privacy (S\&P)}, 2020.

\bibitem[Aghakhani et~al.(2020)Aghakhani, Eisenhofer, Sch{\"o}nherr, Kolossa,
  Holz, Kruegel, and Vigna]{aghakhani-20-venomave}
Hojjat Aghakhani, Thorsten Eisenhofer, Lea Sch{\"o}nherr, Dorothea Kolossa,
  Thorsten Holz, Christopher Kruegel, and Giovanni Vigna.
\newblock {VENOMAVE: Clean-Label Poisoning Against Speech Recognition}.
\newblock \emph{Computing Research Repository (CoRR)}, abs/2010.10682, 2020.

\bibitem[Ahmed et~al.(2020)Ahmed, Kwak, Huh, Kim, Oh, and Kim]{ahmed2020void}
Muhammad~Ejaz Ahmed, Il-Youp Kwak, Jun~Ho Huh, Iljoo Kim, Taekkyung Oh, and
  Hyoungshick Kim.
\newblock {Void: A fast and light voice liveness detection system}.
\newblock In \emph{USENIX Security Symposium}, 2020.

\bibitem[Ai and Ling(2020)]{ai2020neural}
Yang Ai and Zhen-Hua Ling.
\newblock {A Neural Vocoder With Hierarchical Generation of Amplitude and Phase
  Spectra for Statistical Parametric Speech Synthesis}.
\newblock \emph{IEEE/ACM Transactions on Audio, Speech, and Language
  Processing}, 2020.

\bibitem[Alzantot et~al.(2017)Alzantot, Balaji, and
  Srivastava]{alzantot-18-did}
Moustafa Alzantot, Bharathan Balaji, and Mani Srivastava.
\newblock {Did you hear that? Adversarial Examples Against Automatic Speech
  Recognition}.
\newblock In \emph{Advances in Neural Information Processing Systems
  (NeurIPS)}, 2017.

\bibitem[Ardila et~al.(2020)Ardila, Branson, Davis, Henretty, Kohler, Meyer,
  Morais, Saunders, Tyers, and Weber]{ardila2019common}
Rosana Ardila, Megan Branson, Kelly Davis, Michael Henretty, Michael Kohler,
  Josh Meyer, Reuben Morais, Lindsay Saunders, Francis~M Tyers, and Gregor
  Weber.
\newblock {Common Voice: A Massively-Multilingual Speech Corpus}.
\newblock In \emph{Language Resources and Evaluation Conference}, 2020.

\bibitem[Arik et~al.(2017)Arik, Diamos, Gibiansky, Miller, Peng, Ping, Raiman,
  and Zhou]{arik2017deep2}
Sercan Arik, Gregory Diamos, Andrew Gibiansky, John Miller, Kainan Peng, Wei
  Ping, Jonathan Raiman, and Yanqi Zhou.
\newblock {Deep Voice 2: Multi-Speaker Neural Text-to-Speech}.
\newblock In \emph{Advances in Neural Information Processing Systems
  (NeurIPS)}, 2017.

\bibitem[Ar{\i}k et~al.(2017)Ar{\i}k, Chrzanowski, Coates, Diamos, Gibiansky,
  Kang, Li, Miller, Ng, Raiman, et~al.]{arik2017deep}
Sercan~{\"O} Ar{\i}k, Mike Chrzanowski, Adam Coates, Gregory Diamos, Andrew
  Gibiansky, Yongguo Kang, Xian Li, John Miller, Andrew Ng, Jonathan Raiman,
  et~al.
\newblock {Deep Voice: Real-Time Neural Text-to-Speech}.
\newblock In \emph{International Conference on Machine Learning (ICML)}, 2017.

\bibitem[Barker et~al.(2016)]{barker2016guideline}
Elaine Barker et~al.
\newblock {Guideline for Using Cryptographic Standards in the Federal
  Government: Cryptographic Mechanisms}.
\newblock \emph{NIST special publication}, 2016.

\bibitem[Bi{\'n}kowski et~al.(2020)Bi{\'n}kowski, Donahue, Dieleman, Clark,
  Elsen, Casagrande, Cobo, and Simonyan]{binkowski2019high}
Miko{\l}aj Bi{\'n}kowski, Jeff Donahue, Sander Dieleman, Aidan Clark, Erich
  Elsen, Norman Casagrande, Luis~C Cobo, and Karen Simonyan.
\newblock {High Fidelity Speech Synthesis with Adversarial Networks}.
\newblock In \emph{International Conference on Learning Representations
  (ICLR)}, 2020.

\bibitem[Burgess(2020)]{burgess2020telegram}
Matt Burgess.
\newblock {Telegram Still Hasn’t Removed an AI Bot That’s Abusing Women}.
\newblock \emph{Wired}, 2020.

\bibitem[Carlini and Farid(2020)]{carlini2020evading}
Nicholas Carlini and Hany Farid.
\newblock {Evading DeepFake-Image Detectors with White-and Black-Box Attacks}.
\newblock In \emph{IEEE Conference on Computer Vision and Pattern Recognition
  (CVPR)}, 2020.

\bibitem[Carlini and Wagner(2018)]{carlini-18-audio}
Nicholas Carlini and David Wagner.
\newblock {Audio Adversarial Examples: Targeted Attacks on Speech-to-Text}.
\newblock In \emph{IEEE Deep Learning and Security Workshop (DLS)}, 2018.

\bibitem[Carlini et~al.(2019)Carlini, Athalye, Papernot, Brendel, Rauber,
  Tsipras, Goodfellow, and Madry]{carlini-19-evaluating}
Nicholas Carlini, Anish Athalye, Nicolas Papernot, Wieland Brendel, Jonas
  Rauber, Dimitris Tsipras, Ian Goodfellow, and Aleksander Madry.
\newblock {On Evaluating Adversarial Robustness}.
\newblock \emph{Computing Research Repository (CoRR)}, abs/1902.06705, 2019.

\bibitem[Chen et~al.(2020)Chen, Zhang, Zen, Weiss, Norouzi, and
  Chan]{chen2020wavegrad}
Nanxin Chen, Yu~Zhang, Heiga Zen, Ron~J Weiss, Mohammad Norouzi, and William
  Chan.
\newblock {WaveGrad: Estimating Gradients for Waveform Generation}.
\newblock In \emph{International Conference on Learning Representations
  (ICLR)}, 2020.

\bibitem[Chintha et~al.(2020)Chintha, Thai, Sohrawardi, Bhatt, Hickerson,
  Wright, and Ptucha]{chintha2020videodeepfake}
Akash Chintha, Bao Thai, Saniat~Javid Sohrawardi, Kartavya Bhatt, Andrea
  Hickerson, Matthew Wright, and Raymond Ptucha.
\newblock {Recurrent Convolutional Structures for Audio Spoof and Video
  DeepFake Detection}.
\newblock \emph{IEEE Journal of Selected Topics in Signal Processing}, 2020.

\bibitem[Dolhansky et~al.(2020)Dolhansky, Bitton, Pflaum, Lu, Howes, Wang, and
  Ferrer]{dolhansky2020deepfake}
Brian Dolhansky, Joanna Bitton, Ben Pflaum, Jikuo Lu, Russ Howes, Menglin Wang,
  and Cristian~Canton Ferrer.
\newblock {The DeepFake Detection Challenge (DFDC) Dataset}, 2020.

\bibitem[Donahue et~al.(2019)Donahue, McAuley, and
  Puckette]{donahue2018adversarial}
Chris Donahue, Julian McAuley, and Miller Puckette.
\newblock {Adversarial Audio Synthesis}.
\newblock In \emph{International Conference on Learning Representations
  (ICLR)}, 2019.

\bibitem[Donahue et~al.(2021)Donahue, Dieleman, Bi{\'n}kowski, Elsen, and
  Simonyan]{donahue2020end}
Jeff Donahue, Sander Dieleman, Miko{\l}aj Bi{\'n}kowski, Erich Elsen, and Karen
  Simonyan.
\newblock {End-to-End Adversarial Text-to-Speech}.
\newblock In \emph{International Conference on Learning Representations
  (ICLR)}, 2021.

\bibitem[Durall et~al.(2020)Durall, Keuper, and Keuper]{durall2020watch}
Ricard Durall, Margret Keuper, and Janis Keuper.
\newblock {Watch your Up-Convolution: CNN Based Generative Deep Neural Networks
  are Failing to Reproduce Spectral Distributions}.
\newblock In \emph{IEEE Conference on Computer Vision and Pattern Recognition
  (CVPR)}, 2020.

\bibitem[Esteban et~al.(2018)Esteban, Hyland, and R{\"a}tsch]{esteban2017real}
Crist{\'o}bal Esteban, Stephanie~L Hyland, and Gunnar R{\"a}tsch.
\newblock {Real-Valued (Medical) Time Series Generation with Recurrent
  Conditional GANs}.
\newblock In \emph{International Conference on Learning Representations
  (ICLR)}, 2018.

\bibitem[Fan et~al.(2014)Fan, Qian, Xie, and Soong]{fan2014tts}
Yuchen Fan, Yao Qian, Feng-Long Xie, and Frank~K Soong.
\newblock {TTS Synthesis with Bidirectional LSTM Based Recurrent Neural
  Networks}.
\newblock In \emph{International Conference on Acoustics, Speech and Signal
  Processing (ICASSP)}, 2014.

\bibitem[Franceschi-Bicchierai(2020)]{fran2020listen}
Lorenzo Franceschi-Bicchierai.
\newblock {Listen to This Deepfake Audio Impersonating a CEO in Brazen Fraud
  Attempt}.
\newblock \emph{Motherboard}, 2020.

\bibitem[Frank et~al.(2020)Frank, Eisenhofer, Sch{\"o}nherr, Fischer, Kolossa,
  and Holz]{frank2020leveraging}
Joel Frank, Thorsten Eisenhofer, Lea Sch{\"o}nherr, Asja Fischer, Dorothea
  Kolossa, and Thorsten Holz.
\newblock {Leveraging Frequency Analysis for Deep Fake Image Recognition}.
\newblock In \emph{International Conference on Machine Learning (ICML)}, 2020.

\bibitem[Frid-Adar et~al.(2018)Frid-Adar, Diamant, Klang, Amitai, Goldberger,
  and Greenspan]{frid2018gan}
Maayan Frid-Adar, Idit Diamant, Eyal Klang, Michal Amitai, Jacob Goldberger,
  and Hayit Greenspan.
\newblock {GAN-Based synthetic Medical Image Augmentation for Increased CNN
  Performance in Liver Lesion Classification}.
\newblock \emph{Neurocomputing}, 2018.

\bibitem[Goodfellow et~al.(2014)Goodfellow, Pouget-Abadie, Mirza, Xu,
  Warde-Farley, Ozair, Courville, and Bengio]{goodfellow2014generative}
Ian~J Goodfellow, Jean Pouget-Abadie, Mehdi Mirza, Bing Xu, David Warde-Farley,
  Sherjil Ozair, Aaron Courville, and Yoshua Bengio.
\newblock {Generative Adversarial Networks}.
\newblock In \emph{Advances in Neural Information Processing Systems
  (NeurIPS)}, 2014.

\bibitem[Gulati et~al.(2020)Gulati, Qin, Chiu, Parmar, Zhang, Yu, Han, Wang,
  Zhang, Wu, et~al.]{gulati2020conformer}
Anmol Gulati, James Qin, Chung-Cheng Chiu, Niki Parmar, Yu~Zhang, Jiahui Yu,
  Wei Han, Shibo Wang, Zhengdong Zhang, Yonghui Wu, et~al.
\newblock {Conformer: Convolution-Augmented Transformer for Speech
  Recognition}.
\newblock In \emph{Proceedings of Interspeech (INTERSPEECH)}, 2020.

\bibitem[Hao(2019)]{hao2019deepfake}
Karen Hao.
\newblock {The Biggest Threat of Deepfakes isn’t the Deepfakes Themselves}.
\newblock \emph{MIT Technology Review}, 2019.

\bibitem[Hayashi(2020)]{githubPWG}
Tomoki Hayashi.
\newblock {Parallel WaveGAN (+ MelGAN \& Multi-band MelGAN) implementation with
  Pytorch}.
\newblock \url{https://github.com/kan-bayashi/ParallelWaveGAN}, 2020.

\bibitem[Hayashi et~al.(2020)Hayashi, Yamamoto, Inoue, Yoshimura, Watanabe,
  Toda, Takeda, Zhang, and Tan]{hayashi2020espnet}
Tomoki Hayashi, Ryuichi Yamamoto, Katsuki Inoue, Takenori Yoshimura, Shinji
  Watanabe, Tomoki Toda, Kazuya Takeda, Yu~Zhang, and Xu~Tan.
\newblock {Espnet-TTS}: Unified, reproducible, and integratable open source
  end-to-end text-to-speech toolkit.
\newblock In \emph{International Conference on Acoustics, Speech and Signal
  Processing (ICASSP)}, 2020.

\bibitem[Huang et~al.(1979)Huang, Yang, and Tang]{huang-79-fast}
Thomas Huang, GJTGY Yang, and Greory Tang.
\newblock A fast two-dimensional median filtering algorithm.
\newblock \emph{IEEE transactions on acoustics, speech, and signal processing},
  27\penalty0 (1):\penalty0 13--18, 1979.

\bibitem[Inaguma et~al.(2020)Inaguma, Kiyono, Duh, Karita, Yalta, Hayashi, and
  Watanabe]{inaguma-etal-2020-espnet}
Hirofumi Inaguma, Shun Kiyono, Kevin Duh, Shigeki Karita, Nelson Yalta, Tomoki
  Hayashi, and Shinji Watanabe.
\newblock {ESP}net-{ST}: All-in-one speech translation toolkit.
\newblock In \emph{Proceedings of the 58th Annual Meeting of the Association
  for Computational Linguistics: System Demonstrations}, 2020.

\bibitem[Ito and Johnson(2017)]{ljspeech17}
Keith Ito and Linda Johnson.
\newblock {The LJ Speech Dataset}.
\newblock \url{https://keithito.com/LJ-Speech-Dataset/}, 2017.

\bibitem[Jung et~al.(2020)Jung, Kim, Shim, Kim, and Yu]{jung2020improved}
Jee-weon Jung, Seung-bin Kim, Hye-jin Shim, Ju-ho Kim, and Ha-Jin Yu.
\newblock {Improved RawNet with Feature Map Scaling for Text-independent
  Speaker Verification using Raw Waveforms}.
\newblock \emph{Proceedings of Interspeech (INTERSPEECH)}, 2020.

\bibitem[Kalchbrenner et~al.(2018)Kalchbrenner, Elsen, Simonyan, Noury,
  Casagrande, Lockhart, Stimberg, Oord, Dieleman, and
  Kavukcuoglu]{kalchbrenner2018efficient}
Nal Kalchbrenner, Erich Elsen, Karen Simonyan, Seb Noury, Norman Casagrande,
  Edward Lockhart, Florian Stimberg, Aaron Oord, Sander Dieleman, and Koray
  Kavukcuoglu.
\newblock {Efficient Neural Audio Synthesis}.
\newblock In \emph{International Conference on Machine Learning (ICML)}, 2018.

\bibitem[Kamble et~al.(2018)Kamble, Tak, and Patil]{kamble2018effectiveness}
Madhu~R Kamble, Hemlata Tak, and Hemant~A Patil.
\newblock {Effectiveness of Speech Demodulation-Based Features for Replay
  Detection.}
\newblock In \emph{Proceedings of Interspeech (INTERSPEECH)}, 2018.

\bibitem[Kerckhoffs(1883)]{kerckhoffs1883cryptographie}
Auguste Kerckhoffs.
\newblock \emph{{La cryptographie militaire, ou, Des chiffres usit{\'e}s en
  temps de guerre: avec un nouveau proc{\'e}d{\'e} de d{\'e}chiffrement
  applicable aux syst{\`e}mes {\`a} double clef}}.
\newblock Librairie militaire de L. Baudoin, 1883.

\bibitem[Killoran et~al.(2017)Killoran, Lee, Delong, Duvenaud, and
  Frey]{killoran2017generating}
Nathan Killoran, Leo~J Lee, Andrew Delong, David Duvenaud, and Brendan~J Frey.
\newblock {Generating and designing DNA with deep generative models}.
\newblock \emph{arXiv preprint arXiv:1712.06148}, 2017.

\bibitem[Kim et~al.(2019)Kim, Lee, Song, Kim, and Yoon]{kim2018flowavenet}
Sungwon Kim, Sang-Gil Lee, Jongyoon Song, Jaehyeon Kim, and Sungroh Yoon.
\newblock {FloWaveNet: A Generative Flow for Raw Audio}.
\newblock In \emph{International Conference on Machine Learning (ICML)}, 2019.

\bibitem[Kingma and Ba(2015)]{kingma2014adam}
Diederik~P Kingma and Jimmy Ba.
\newblock {Adam: A method for stochastic optimization}.
\newblock In \emph{International Conference on Learning Representations
  (ICLR)}, 2015.

\bibitem[Kingma and Dhariwal(2019)]{kingma2018glow}
Diederik~P Kingma and Prafulla Dhariwal.
\newblock {Glow: Generative Flow with Invertible 1x1 Convolutions}.
\newblock In \emph{Advances in Neural Information Processing Systems
  (NeurIPS)}, 2019.

\bibitem[Kingma et~al.(2016)Kingma, Salimans, Jozefowicz, Chen, Sutskever, and
  Welling]{kingma2016improving}
Diederik~P Kingma, Tim Salimans, Rafal Jozefowicz, Xi~Chen, Ilya Sutskever, and
  Max Welling.
\newblock {Improving Variational Inference with Inverse Autoregressive Flow}.
\newblock In \emph{International Conference on Learning Representations (ICLR)
  - Workshop track}, 2016.

\bibitem[Kinnunen et~al.(2017)Kinnunen, Sahidullah, Falcone, Costantini,
  Hautamäki, Thomsen, Sarkar, Tan, Delgado, Todisco, Evans, Hautamäki, and
  Lee]{kinnunen2017reddots}
Tomi Kinnunen, Md~Sahidullah, Mauro Falcone, Luca Costantini, Rosa~González
  Hautamäki, Dennis Thomsen, Achintya Sarkar, Zheng-Hua Tan, Héctor Delgado,
  Massimiliano Todisco, Nicholas Evans, Ville Hautamäki, and Kong~Aik Lee.
\newblock {RedDots replayed: A new replay spoofing attack corpus for
  text-dependent speaker verification research}.
\newblock In \emph{International Conference on Acoustics, Speech and Signal
  Processing (ICASSP)}, 2017.

\bibitem[Kocher et~al.(2019)Kocher, Horn, Fogh, , Genkin, Gruss, Haas, Hamburg,
  Lipp, Mangard, Prescher, Schwarz, and Yarom]{Kocher2018spectre}
Paul Kocher, Jann Horn, Anders Fogh, , Daniel Genkin, Daniel Gruss, Werner
  Haas, Mike Hamburg, Moritz Lipp, Stefan Mangard, Thomas Prescher, Michael
  Schwarz, and Yuval Yarom.
\newblock Spectre attacks: Exploiting speculative execution.
\newblock In \emph{40th IEEE Symposium on Security and Privacy (S\&P'19)},
  2019.

\bibitem[Kong et~al.(2020{\natexlab{a}})Kong, Kim, and Bae]{githubHiFi}
Jungil Kong, Jaehyeon Kim, and Jaekyoung Bae.
\newblock {Hifi-GAN: Generative Adversarial Networks for Efficient and High
  Fidelity Speech Synthesis}.
\newblock \url{https://github.com/jik876/hifi-gan}, 2020{\natexlab{a}}.

\bibitem[Kong et~al.(2020{\natexlab{b}})Kong, Kim, and Bae]{kong2020hifi}
Jungil Kong, Jaehyeon Kim, and Jaekyoung Bae.
\newblock {Hifi-GAN: Generative Adversarial Networks for Efficient and High
  Fidelity Speech Synthesis}.
\newblock In \emph{Advances in Neural Information Processing Systems
  (NeurIPS)}, 2020{\natexlab{b}}.

\bibitem[Kong et~al.(2021)Kong, Ping, Huang, Zhao, and
  Catanzaro]{kong2020diffwave}
Zhifeng Kong, Wei Ping, Jiaji Huang, Kexin Zhao, and Bryan Catanzaro.
\newblock {DiffWave: A Versatile Diffusion Model for Audio Synthesis}.
\newblock In \emph{International Conference on Learning Representations
  (ICLR)}, 2021.

\bibitem[Kumar et~al.(2019)Kumar, Kumar, de~Boissiere, Gestin, Teoh, Sotelo,
  de~Br{\'e}bisson, Bengio, and Courville]{kumar2019melgan}
Kundan Kumar, Rithesh Kumar, Thibault de~Boissiere, Lucas Gestin, Wei~Zhen
  Teoh, Jose Sotelo, Alexandre de~Br{\'e}bisson, Yoshua Bengio, and Aaron
  Courville.
\newblock {MelGAN: Generative Adversarial Networks for Conditional Waveform
  Synthesis}.
\newblock In \emph{Advances in Neural Information Processing Systems
  (NeurIPS)}, 2019.

\bibitem[Lab(2021)]{dfrlab2021navalny}
The Atlantic Council’s Digital Forensic~Research Lab.
\newblock {Inauthentic Instagram accounts with synthetic faces target Navalny
  protests}.
\newblock \emph{Medium}, 2021.

\bibitem[Lai et~al.(2019)Lai, Abad, Richmond, Yamagishi, Dehak, and
  King]{lai2019attentive}
Cheng-I Lai, Alberto Abad, Korin Richmond, Junichi Yamagishi, Najim Dehak, and
  Simon King.
\newblock {Attentive Filtering Networks for Audio Replay Attack Detection}.
\newblock In \emph{International Conference on Acoustics, Speech and Signal
  Processing (ICASSP)}, 2019.

\bibitem[Lavrentyeva et~al.(2017)Lavrentyeva, Novoselov, Malykh, Kozlov,
  Kudashev, and Shchemelinin]{lavrentyeva2017audio}
Galina Lavrentyeva, Sergey Novoselov, Egor Malykh, Alexander Kozlov, Oleg
  Kudashev, and Vadim Shchemelinin.
\newblock {Audio Replay Attack Detection with Deep Learning Frameworks.}
\newblock In \emph{Proceedings of Interspeech (INTERSPEECH)}, 2017.

\bibitem[Li et~al.(2021)Li, Shi, Zhang, Subramanian, Chang, Kamo, Hira,
  Hayashi, Boeddeker, Chen, and Watanabe]{li2020espnet}
Chenda Li, Jing Shi, Wangyou Zhang, Aswin~Shanmugam Subramanian, Xuankai Chang,
  Naoyuki Kamo, Moto Hira, Tomoki Hayashi, Christoph Boeddeker, Zhuo Chen, and
  Shinji Watanabe.
\newblock {ESPnet-SE}: End-to-end speech enhancement and separation toolkit
  designed for {ASR} integration.
\newblock In \emph{Proceedings of IEEE Spoken Language Technology Workshop
  (SLT)}, 2021.

\bibitem[Li and Lyu(2018)]{li2018exposing}
Yuezun Li and Siwei Lyu.
\newblock {Exposing DeepFake Videos by Detecting Face Warping Artifacts}.
\newblock \emph{arXiv preprint arXiv:1811.00656}, 2018.

\bibitem[Li et~al.(2020)Li, Yang, Sun, Qi, and Lyu]{li2020celeb}
Yuezun Li, Xin Yang, Pu~Sun, Honggang Qi, and Siwei Lyu.
\newblock {Celeb-DF: A New Dataset for DeepFake Forensics}.
\newblock In \emph{Proceedings of the IEEE/CVF Conference on Computer Vision
  and Pattern Recognition}, 2020.

\bibitem[Lipp et~al.(2018)Lipp, Schwarz, Gruss, Prescher, Haas, Fogh, Horn,
  Mangard, Kocher, Genkin, Yarom, and Hamburg]{Lipp2018meltdown}
Moritz Lipp, Michael Schwarz, Daniel Gruss, Thomas Prescher, Werner Haas,
  Anders Fogh, Jann Horn, Stefan Mangard, Paul Kocher, Daniel Genkin, Yuval
  Yarom, and Mike Hamburg.
\newblock Meltdown: Reading kernel memory from user space.
\newblock In \emph{27th {USENIX} Security Symposium ({USENIX} Security 18)},
  2018.

\bibitem[Marra et~al.(2018)Marra, Gragnaniello, Cozzolino, and
  Verdoliva]{marra2018detection}
Francesco Marra, Diego Gragnaniello, Davide Cozzolino, and Luisa Verdoliva.
\newblock {Detection of GAN-Generated Fake Images over Social Networks}.
\newblock In \emph{IEEE Conference on Multimedia Information Processing and
  Retrieval (MIPR)}, 2018.

\bibitem[Marra et~al.(2019)Marra, Gragnaniello, Verdoliva, and
  Poggi]{marra2019gans}
Francesco Marra, Diego Gragnaniello, Luisa Verdoliva, and Giovanni Poggi.
\newblock {Do GANs Leave Artificial Gingerprints?}
\newblock In \emph{IEEE Conference on Multimedia Information Processing and
  Retrieval (MIPR)}, 2019.

\bibitem[McCarty(2003)]{mccarty2003honeynet}
Bill McCarty.
\newblock {The honeynet arms race}.
\newblock \emph{IEEE Security \& Privacy}, 2003.

\bibitem[McCloskey and Albright(2018)]{mccloskey2018detecting}
Scott McCloskey and Michael Albright.
\newblock {Detecting GAN-Generated Imagery Using Color Cues}.
\newblock \emph{arXiv preprint arXiv:1812.08247}, 2018.

\bibitem[Mehri et~al.(2016)Mehri, Kumar, Gulrajani, Kumar, Jain, Sotelo,
  Courville, and Bengio]{mehri2016samplernn}
Soroush Mehri, Kundan Kumar, Ishaan Gulrajani, Rithesh Kumar, Shubham Jain,
  Jose Sotelo, Aaron Courville, and Yoshua Bengio.
\newblock {SampleRNN: An Unconditional End-to-End Neural Audio Generation
  Model}.
\newblock \emph{arXiv preprint arXiv:1612.07837}, 2016.

\bibitem[Mo et~al.(2018)Mo, Chen, and Luo]{mo2018fake}
Huaxiao Mo, Bolin Chen, and Weiqi Luo.
\newblock {Fake Faces Identification via Convolutional Neural Network}.
\newblock In \emph{ACM Workshop on Information Hiding and Multimedia Security},
  2018.

\bibitem[Mwai(2021)]{mwai2021tigray}
Peter Mwai.
\newblock {Tigray conflict: The fake UN diplomat and other misleading stories}.
\newblock \emph{BBC Reality Check}, 2021.

\bibitem[Nataraj et~al.(2019)Nataraj, Mohammed, Manjunath, Chandrasekaran,
  Flenner, Bappy, and Roy-Chowdhury]{nataraj2019detecting}
Lakshmanan Nataraj, Tajuddin~Manhar Mohammed, BS~Manjunath, Shivkumar
  Chandrasekaran, Arjuna Flenner, Jawadul~H Bappy, and Amit~K Roy-Chowdhury.
\newblock {Detecting GAN Generated Fake Images Using Co-Occurrence Matrices}.
\newblock \emph{Electronic Imaging}, 2019.

\bibitem[Nautsch et~al.(2021)Nautsch, Wang, Evans, Kinnunen, Vestman, Todisco,
  Delgado, Sahidullah, Yamagishi, and Lee]{nautsch19asvspoof}
Andreas Nautsch, Xin Wang, Nicholas Evans, Tomi~H. Kinnunen, Ville Vestman,
  Massimiliano Todisco, Héctor Delgado, Md~Sahidullah, Junichi Yamagishi, and
  Kong~Aik Lee.
\newblock {ASVspoof 2019: Spoofing Countermeasures for the Detection of
  Synthesized, Converted and Replayed Speech}.
\newblock \emph{IEEE Transactions on Biometrics, Behavior, and Identity
  Science}, 2021.

\bibitem[Oord et~al.(2016)Oord, Dieleman, Zen, Simonyan, Vinyals, Graves,
  Kalchbrenner, Senior, and Kavukcuoglu]{oord2016wavenet}
Aaron van~den Oord, Sander Dieleman, Heiga Zen, Karen Simonyan, Oriol Vinyals,
  Alex Graves, Nal Kalchbrenner, Andrew Senior, and Koray Kavukcuoglu.
\newblock {WaveNet: A Generative Model for Raw Audio}.
\newblock \emph{arXiv preprint arXiv:1609.03499}, 2016.

\bibitem[Oord et~al.(2017)Oord, Vinyals, and Kavukcuoglu]{oord2017neural}
Aaron van~den Oord, Oriol Vinyals, and Koray Kavukcuoglu.
\newblock {Neural Discrete Representation Learning}.
\newblock In \emph{Advances in Neural Information Processing Systems
  (NeurIPS)}, 2017.

\bibitem[Paszke et~al.(2019)Paszke, Gross, Massa, Lerer, Bradbury, Chanan,
  Killeen, Lin, Gimelshein, Antiga, Desmaison, Kopf, Yang, DeVito, Raison,
  Tejani, Chilamkurthy, Steiner, Fang, Bai, and Chintala]{pytorch}
Adam Paszke, Sam Gross, Francisco Massa, Adam Lerer, James Bradbury, Gregory
  Chanan, Trevor Killeen, Zeming Lin, Natalia Gimelshein, Luca Antiga, Alban
  Desmaison, Andreas Kopf, Edward Yang, Zachary DeVito, Martin Raison, Alykhan
  Tejani, Sasank Chilamkurthy, Benoit Steiner, Lu~Fang, Junjie Bai, and Soumith
  Chintala.
\newblock {PyTorch: An Imperative Style, High-Performance Deep Learning
  Library}.
\newblock In \emph{Advances in Neural Information Processing Systems
  (NeurIPS)}, 2019.

\bibitem[Patil et~al.(2017)Patil, Kamble, Patel, and Soni]{patil2017novel}
Hemant~A Patil, Madhu~R Kamble, Tanvina~B Patel, and Meet~H Soni.
\newblock {Novel Variable Length Teager Energy Separation Based Instantaneous
  Frequency Features for Replay Detection.}
\newblock In \emph{Proceedings of Interspeech (INTERSPEECH)}, 2017.

\bibitem[Peng et~al.(2020)Peng, Ping, Song, and Zhao]{peng2020non}
Kainan Peng, Wei Ping, Zhao Song, and Kexin Zhao.
\newblock {Non-Autoregressive Neural Text-to-Speech}.
\newblock In \emph{International Conference on Machine Learning (ICML)}, 2020.

\bibitem[Ping et~al.(2019)Ping, Peng, and Chen]{ping2018clarinet}
Wei Ping, Kainan Peng, and Jitong Chen.
\newblock {Clarinet: Parallel Wave Generation in End-to-End Text-to-Speech}.
\newblock In \emph{International Conference on Learning Representations
  (ICLR)}, 2019.

\bibitem[Ping et~al.(2020)Ping, Peng, Zhao, and Song]{ping2020waveflow}
Wei Ping, Kainan Peng, Kexin Zhao, and Zhao Song.
\newblock {WaveFlow: A Compact Flow-based Model for Raw Audio}.
\newblock In \emph{International Conference on Learning Representations
  (ICLR)}, 2020.

\bibitem[Prabhu(2014)]{prabhu2014window}
KM~Muraleedhara Prabhu.
\newblock \emph{{Window Functions and their Applications in Signal
  Processing}}.
\newblock Taylor \& Francis, 2014.

\bibitem[Prenger et~al.(2018{\natexlab{a}})Prenger, Valle, and
  Catanzaro]{githubTorchhub}
Ryan Prenger, Rafael Valle, and Bryan Catanzaro.
\newblock {WaveGlow: a Flow-based Generative Network for Speech Synthesis}.
\newblock \url{https://pytorch.org/hub/nvidia_deeplearningexamples_waveglow/},
  2018{\natexlab{a}}.

\bibitem[Prenger et~al.(2018{\natexlab{b}})Prenger, Valle, and
  Catanzaro]{githubWaveglow}
Ryan Prenger, Rafael Valle, and Bryan Catanzaro.
\newblock {WaveGlow: a Flow-based Generative Network for Speech Synthesis}.
\newblock \url{https://github.com/NVIDIA/waveglow}, 2018{\natexlab{b}}.

\bibitem[Prenger et~al.(2019)Prenger, Valle, and
  Catanzaro]{prenger2019waveglow}
Ryan Prenger, Rafael Valle, and Bryan Catanzaro.
\newblock {Waveglow: A Flow-based Generative Network for Speech Synthesis}.
\newblock In \emph{International Conference on Acoustics, Speech and Signal
  Processing (ICASSP)}, 2019.

\bibitem[Qian et~al.(2020)Qian, Yin, Sheng, Chen, and Shao]{qian2020thinking}
Yuyang Qian, Guojun Yin, Lu~Sheng, Zixuan Chen, and Jing Shao.
\newblock {Thinking in Frequency: Face Forgery Detection by Mining
  Frequency-Aware Clues}.
\newblock In \emph{European Conference on Computer Vision (ECCV)}, 2020.

\bibitem[Quatieri(2006)]{quatieri2006discrete}
Thomas Quatieri.
\newblock \emph{{Discrete-Time Speech Signal Processing: Principles and
  Practice}}.
\newblock Pearson Education India, 2006.

\bibitem[Rabiner et~al.(2016)Rabiner, Gold, and Yuen]{rabiner2016theory}
Lawrence Rabiner, Bernard Gold, and CK~Yuen.
\newblock \emph{{Theory and Application of Digital Signal Processing}}.
\newblock Prentice-Hall, 2016.

\bibitem[Ren et~al.(2019)Ren, Ruan, Tan, Qin, Zhao, Zhao, and
  Liu]{ren2019fastspeech}
Yi~Ren, Yangjun Ruan, Xu~Tan, Tao Qin, Sheng Zhao, Zhou Zhao, and Tie-Yan Liu.
\newblock {FastSpeech: Fast, Robust and Controllable Text to Speech}.
\newblock In \emph{Advances in Neural Information Processing Systems
  (NeurIPS)}, 2019.

\bibitem[Ren et~al.(2020)Ren, Hu, Tan, Qin, Zhao, Zhao, and
  Liu]{ren2020fastspeech}
Yi~Ren, Chenxu Hu, Xu~Tan, Tao Qin, Sheng Zhao, Zhou Zhao, and Tie-Yan Liu.
\newblock {FastSpeech 2: Fast and High-Quality End-to-End Text to Speech}.
\newblock \emph{arXiv preprint arXiv:2006.04558}, 2020.

\bibitem[Rijmen and Daemen(2001)]{rijmen2001advanced}
Vincent Rijmen and Joan Daemen.
\newblock {Advanced Encryption Standard}.
\newblock \emph{Proceedings of Federal Information Processing Standards
  Publications, National Institute of Standards and Technology}, 2001.

\bibitem[Rossler et~al.(2019)Rossler, Cozzolino, Verdoliva, Riess, Thies, and
  Nie{\ss}ner]{rossler2019faceforensics++}
Andreas Rossler, Davide Cozzolino, Luisa Verdoliva, Christian Riess, Justus
  Thies, and Matthias Nie{\ss}ner.
\newblock {Faceforensics++: Learning to Detect Manipulated Facial Images}.
\newblock In \emph{Proceedings of the IEEE/CVF International Conference on
  Computer Vision}, 2019.

\bibitem[Sahidullah et~al.(2015)Sahidullah, Kinnunen, and
  Hanil{\c{c}}i]{sahidullah2015comparison}
Md~Sahidullah, Tomi Kinnunen, and Cemal Hanil{\c{c}}i.
\newblock {A Comparison of Features for Synthetic Speech Detection}.
\newblock In \emph{Proceedings of Interspeech (INTERSPEECH)}, 2015.

\bibitem[Scarfone et~al.(2008)Scarfone, Jansen, Tracy,
  et~al.]{scarfone2008guide}
Karen Scarfone, Wayne Jansen, Miles Tracy, et~al.
\newblock {Guide to General Server Security}.
\newblock \emph{NIST Special Publication}, 2008.

\bibitem[Scheuermann et~al.(2000)Scheuermann, Schwiderski-Grosche, and
  Struif]{scheuermann2000usability}
Dirk Scheuermann, Scarlet Schwiderski-Grosche, and Bruno Struif.
\newblock \emph{{Usability of Biometrics in Relation to Electronic
  Signatures}}.
\newblock GMD-Forschungszentrum Informationstechnik Sankt Augustin, 2000.

\bibitem[Sch{\"o}nherr et~al.(2017)Sch{\"o}nherr, Zeiler, and
  Kolossa]{schonherr2017lol}
Lea Sch{\"o}nherr, Steffen Zeiler, and Dorothea Kolossa.
\newblock {Spoofing Detection via Simultaneous Verification of Audio-Visual
  Synchronicity and Transcription}.
\newblock In \emph{2017 IEEE Automatic Speech Recognition and Understanding
  Workshop (ASRU)}, 2017.

\bibitem[Sch\"{o}nherr et~al.(2019)Sch\"{o}nherr, Kohls, Zeiler, Holz, and
  Kolossa]{schoenherr-19-psychoacoustics}
Lea Sch\"{o}nherr, Katharina Kohls, Steffen Zeiler, Thorsten Holz, and Dorothea
  Kolossa.
\newblock {Adversarial Attacks Against Automatic Speech Recognition Systems via
  Psychoacoustic Hiding}.
\newblock In \emph{Symposium on Network and Distributed System Security
  (NDSS)}, 2019.

\bibitem[Sch\"{o}nherr et~al.(2020)Sch\"{o}nherr, Eisenhofer, Zeiler, Holz, and
  Kolossa]{schoenherr-20-imperio}
Lea Sch\"{o}nherr, Thorsten Eisenhofer, Steffen Zeiler, Thorsten Holz, and
  Dorothea Kolossa.
\newblock {Imperio: Robust Over-the-Air Adversarial Examples for Automatic
  Speech Recognition Systems}.
\newblock In \emph{Annual Computer Security Applications Conference (ACSAC)},
  2020.

\bibitem[Shannon(1949)]{shannon1949communication}
Claude~E Shannon.
\newblock {Communication Theory of Secrecy Systems}.
\newblock \emph{The Bell system technical journal}, 1949.

\bibitem[Shen et~al.(2018)Shen, Pang, Weiss, Schuster, Jaitly, Yang, Chen,
  Zhang, Wang, Skerrv-Ryan, et~al.]{shen2018natural}
Jonathan Shen, Ruoming Pang, Ron~J Weiss, Mike Schuster, Navdeep Jaitly,
  Zongheng Yang, Zhifeng Chen, Yu~Zhang, Yuxuan Wang, Rj~Skerrv-Ryan, et~al.
\newblock {Natural TTS Synthesis by Conditioning WaveNet on Mel Spectrogram
  Predictions}.
\newblock In \emph{International Conference on Acoustics, Speech and Signal
  Processing (ICASSP)}, 2018.

\bibitem[Sommer and Paxson(2010)]{sommer2010outside}
Robin Sommer and Vern Paxson.
\newblock {Outside the Closed World: On Using Machine Learning for Network
  Intrusion Detection}.
\newblock In \emph{IEEE Symposium on Security and Privacy (S\&P)}, 2010.

\bibitem[Sonobe et~al.(2017)Sonobe, Takamichi, and Saruwatari]{sonobe2017jsut}
Ryosuke Sonobe, Shinnosuke Takamichi, and Hiroshi Saruwatari.
\newblock {JSUT Corpus: Free Large-Scale Japanese Speech Corpus for End-to-End
  Speech Synthesis}.
\newblock \emph{arXiv preprint arXiv:1711.00354}, 2017.

\bibitem[Sotelo et~al.(2017)Sotelo, Mehri, Kumar, Santos, Kastner, Courville,
  and Bengio]{sotelo2017char2wav}
Jose Sotelo, Soroush Mehri, Kundan Kumar, Joao~Felipe Santos, Kyle Kastner,
  Aaron Courville, and Yoshua Bengio.
\newblock {Char2wav: End-to-End Speech Synthesis}.
\newblock In \emph{International Conference on Learning Representations (ICLR)
  Workshop Track}, 2017.

\bibitem[Stupp(2019)]{stupp2019ceo}
Catherine Stupp.
\newblock {Fraudsters Used AI to Mimic CEO’s Voice in Unusual Cybercrime
  Case}.
\newblock \emph{The Wall Street Journal}, 2019.

\bibitem[Taigman et~al.(2017)Taigman, Wolf, Polyak, and
  Nachmani]{taigman2017voiceloop}
Yaniv Taigman, Lior Wolf, Adam Polyak, and Eliya Nachmani.
\newblock {VoiceLoop: Voice Fitting and Synthesis via a Phonological Loop}.
\newblock In \emph{International Conference on Learning Representations
  (ICLR)}, 2017.

\bibitem[Tak et~al.(2021)Tak, Patino, Todisco, Nautsch, Evans, and
  Larcher]{tak2021rawnet2}
Hemlata Tak, Jose Patino, Massimiliano Todisco, Andreas Nautsch, Nicholas
  Evans, and Anthony Larcher.
\newblock {End-to-End anti-spoofing with RawNet2}.
\newblock In \emph{International Conference on Acoustics, Speech and Signal
  Processing (ICASSP)}, 2021.

\bibitem[Tariq et~al.(2019)Tariq, Lee, Kim, Shin, and Woo]{tariq2019gan}
Shahroz Tariq, Sangyup Lee, Hoyoung Kim, Youjin Shin, and Simon~S Woo.
\newblock {GAN is a Friend or Foe? A Framework to Detect Various Fake Face
  Images}.
\newblock In \emph{ACM/SIGAPP Symposium on Applied Computing}, 2019.

\bibitem[Todisco et~al.(2019)Todisco, Wang, Vestman, Sahidullah, Delgado,
  Nautsch, Yamagishi, Evans, Kinnunen, and Lee]{todisco2019asvspoof}
Massimiliano Todisco, Xin Wang, Ville Vestman, Md~Sahidullah, Hector Delgado,
  Andreas Nautsch, Junichi Yamagishi, Nicholas Evans, Tomi Kinnunen, and
  Kong~Aik Lee.
\newblock {ASVspoof 2019: Future Horizons in Spoofed and Fake Audio Detection}.
\newblock \emph{Computing Research Repository (CoRR)}, abs/1904.05441, 2019.

\bibitem[Tokuda et~al.(2000)Tokuda, Yoshimura, Masuko, Kobayashi, and
  Kitamura]{tokuda2000speech}
Keiichi Tokuda, Takayoshi Yoshimura, Takashi Masuko, Takao Kobayashi, and
  Tadashi Kitamura.
\newblock {Speech Parameter Generation Algorithms for HMM-Based Speech
  Synthesis}.
\newblock In \emph{International Conference on Acoustics, Speech and Signal
  Processing (ICASSP)}, 2000.

\bibitem[Tom et~al.(2018)Tom, Jain, and Dey]{tom2018end}
Francis Tom, Mohit Jain, and Prasenjit Dey.
\newblock {End-To-End Audio Replay Attack Detection Using Deep Convolutional
  Networks with Attention.}
\newblock In \emph{Proceedings of Interspeech (INTERSPEECH)}, 2018.

\bibitem[Valle et~al.(2018)Valle, Cai, and Doshi]{valle2018tequilagan}
Rafael Valle, Wilson Cai, and Anish Doshi.
\newblock {TequilaGAN: How to Easily Identify GAN Samples}.
\newblock \emph{arXiv preprint arXiv:1807.04919}, 2018.

\bibitem[Wang et~al.(2020)Wang, Wang, Zhang, Owens, and Efros]{wang2019cnn}
Sheng-Yu Wang, Oliver Wang, Richard Zhang, Andrew Owens, and Alexei~A Efros.
\newblock {CNN-generated images are surprisingly easy to spot... for now}.
\newblock In \emph{IEEE Conference on Computer Vision and Pattern Recognition
  (CVPR)}, 2020.

\bibitem[Wang et~al.(2019)Wang, Takaki, and Yamagishi]{wang2019neural}
Xin Wang, Shinji Takaki, and Junichi Yamagishi.
\newblock {Neural Source-Filter Waveform Models for Statistical Parametric
  Speech Synthesis}.
\newblock \emph{IEEE/ACM Transactions on Audio, Speech, and Language
  Processing}, 2019.

\bibitem[Watanabe et~al.(2018)Watanabe, Hori, Karita, Hayashi, Nishitoba, Unno,
  {Enrique Yalta Soplin}, Heymann, Wiesner, Chen, Renduchintala, and
  Ochiai]{watanabe2018espnet}
Shinji Watanabe, Takaaki Hori, Shigeki Karita, Tomoki Hayashi, Jiro Nishitoba,
  Yuya Unno, Nelson {Enrique Yalta Soplin}, Jahn Heymann, Matthew Wiesner,
  Nanxin Chen, Adithya Renduchintala, and Tsubasa Ochiai.
\newblock {ESPnet}: End-to-end speech processing toolkit.
\newblock In \emph{Proceedings of Interspeech (INTERSPEECH)}, 2018.

\bibitem[Willis(2020)]{zero2020policy}
Tim Willis.
\newblock {Project Zero Policy and Disclosure: 2020 Edition}, 2020.
\newblock
  \url{https://googleprojectzero.blogspot.com/2020/01/policy-and-disclosure-2020-edition.html},
  as of \today.

\bibitem[Wu et~al.(2017)Wu, Yamagishi, Kinnunen, Hanilçi, Sahidullah, Sizov,
  Evans, Todisco, and Delgado]{wu2017asvspoof}
Zhizheng Wu, Junichi Yamagishi, Tomi Kinnunen, Cemal Hanilçi, Mohammed
  Sahidullah, Aleksandr Sizov, Nicholas Evans, Massimiliano Todisco, and
  Héctor Delgado.
\newblock {ASVspoof: The Automatic Speaker Verification Spoofing and
  Countermeasures Challenge}.
\newblock \emph{IEEE Journal of Selected Topics in Signal Processing}, 2017.

\bibitem[Xu et~al.(2020)Xu, Venugopalan, and Sundararajan]{xu2020attribution}
Shawn Xu, Subhashini Venugopalan, and Mukund Sundararajan.
\newblock {Attribution in Scale and Space}.
\newblock In \emph{IEEE Conference on Computer Vision and Pattern Recognition
  (CVPR)}, 2020.

\bibitem[Yamamoto et~al.(2020)Yamamoto, Song, and Kim]{yamamoto2020pwg}
Ryuichi Yamamoto, Eunwoo Song, and Jae-Min Kim.
\newblock {Parallel WaveGAN: A Fast Waveform Generation Model Based on
  Generative Adversarial Networks with Multi-Resolution Spectrogram}.
\newblock In \emph{International Conference on Acoustics, Speech and Signal
  Processing (ICASSP)}, 2020.

\bibitem[Yang et~al.(2021)Yang, Yang, Liu, Fang, Chen, and
  Xie]{yang2021mbmelgan}
Geng Yang, Shan Yang, Kai Liu, Peng Fang, Wei Chen, and Lei Xie.
\newblock {Multi-Band Melgan: Faster Waveform Generation For High-Quality
  Text-To-Speech}.
\newblock In \emph{Proceedings of IEEE Spoken Language Technology Workshop
  (SLT)}, 2021.

\bibitem[Yoshimura et~al.(1999)Yoshimura, Tokuda, Masuko, Kobayashi, and
  Kitamura]{yoshimura1999simultaneous}
Takayoshi Yoshimura, Keiichi Tokuda, Takashi Masuko, Takao Kobayashi, and
  Tadashi Kitamura.
\newblock {Simultaneous Modeling of Spectrum, Pitch and Duration in HMM-Based
  Speech Synthesis}.
\newblock In \emph{Sixth European Conference on Speech Communication and
  Technology}, 1999.

\bibitem[Yu et~al.(2019)Yu, Davis, and Fritz]{yu2019attributing}
Ning Yu, Larry~S Davis, and Mario Fritz.
\newblock {Attributing Fake Images to GANs: Learning and Analyzing GAN
  Fingerprints}.
\newblock In \emph{IEEE International Conference on Computer Vision (ICCV)},
  2019.

\bibitem[Yuan et~al.(2018)Yuan, Chen, Zhao, Long, Liu, Chen, Zhang, Huang,
  Wang, and Gunter]{yuan-18-commandersong}
Xuejing Yuan, Yuxuan Chen, Yue Zhao, Yunhui Long, Xiaokang Liu, Kai Chen,
  Shengzhi Zhang, Heqing Huang, Xiaofeng Wang, and Carl~A. Gunter.
\newblock {CommanderSong: A Systematic Approach for Practical Adversarial Voice
  Recognition}.
\newblock In \emph{USENIX Security Symposium}, 2018.

\bibitem[Ze et~al.(2013)Ze, Senior, and Schuster]{ze2013statistical}
Heiga Ze, Andrew Senior, and Mike Schuster.
\newblock {Statistical Parametric Speech Synthesis Using Deep Neural Networks}.
\newblock In \emph{International Conference on Acoustics, Speech and Signal
  Processing (ICASSP)}, 2013.

\bibitem[Zen et~al.(2009)Zen, Tokuda, and Black]{zen2009statistical}
Heiga Zen, Keiichi Tokuda, and Alan~W Black.
\newblock {Statistical Parametric Speech Synthesis}.
\newblock \emph{Speech Communication}, 2009.

\bibitem[Zhang et~al.(2016)Zhang, Tan, Yang, and Chen]{zhang2016voicelive}
Linghan Zhang, Sheng Tan, Jie Yang, and Yingying Chen.
\newblock {Voicelive: A Phoneme Localization Based Liveness Detection for Voice
  Authentication on Smartphones}.
\newblock In \emph{ACM Conference on Computer and Communications Security
  (CCS)}, 2016.

\bibitem[Zhang et~al.(2019)Zhang, Karaman, and Chang]{zhang2019detecting}
Xu~Zhang, Svebor Karaman, and Shih-Fu Chang.
\newblock {Detecting and Simulating Artifacts in GAN Fake Images}.
\newblock In \emph{2019 IEEE International Workshop on Information Forensics
  and Security (WIFS)}, 2019.

\bibitem[Zwicker and Fastl(2007)]{zwicker-07-psychoacoustics}
Eberhard Zwicker and Hugo Fastl.
\newblock \emph{{Psychoacoustics: Facts and Models}}.
\newblock Springer, Heidelberg, Germany, 2 edition, 2007.

\end{thebibliography}
\clearpage

\section*{Supplementary Material}
\appendix

In this supplementary material, we provide an extended discussion on security research, a note on licensing generative models, our training details, the results on \mfcc{} data, and full-size spectrogram and attribution plots.
Additionally, we provide a visual representation of the filterbanks.

\section{A note on releasing security research}

One might wonder if releasing research into detecting Deepfakes might negatively affect the detection ''arms race``.
That is a long-standing debate in the security community. The overall consensus is that ''security through obscurity`` does not work.
This is often echoed in best security practices, for example, published by the  National Institute of Standards and Technology (NIST)~\citep{scarfone2008guide}.
Intuitively, withholding information from the research community is more harmful since attackers will eventually adapt to any defense one deploys anyway.
Thus, contributing to the invention of new systems is more helpful in an ever-changing environment~\citep{mccarty2003honeynet}

The debate dates back to at least the 19th century where the cryptographer Auguste Kerckhoffs introduced Kerckhoffs's principle~\citep{kerckhoffs1883cryptographie}.
The principle states that an encryption scheme should still work if an adversary knows everything about the system but a secret passphrase.
Similar thought would later be formulated by Claude Shannon~\citep{shannon1949communication}.

A typical example is the advanced encryption standard (AES).
The algorithms entire specification and inner workings can be found in the standardization~\citep{rijmen2001advanced}.
Yet, it is considered unbreakable as long as the password used for the encryption is not revealed.
AES is also the only algorithm used to encrypt US government documents~\citep{barker2016guideline}.
The principle also found adoption in the machine learning community, where adversarial defense papers are now advised to evaluate against so-called white box attackers~\citep{carlini-19-evaluating}, i.e., attackers which know the inner workings of the system and actively try to avoid it.

While complete openness is not possible, the greater security community has adopted similar practices.
For example, so-called attack papers are regularly published at security venues.
The underlying motivation being, that before one can protect systems, one has to understand how to attack them.
Prominent examples are the Meltdown~\citep{Lipp2018meltdown} and Spectre~\citep{Kocher2018spectre} vulnerabilities which showed that certain instructions in CPUs could be used for unauthorized access.

Similar patterns are also used in the industry.
Google's project zero team regularly analyses and finds critical vulnerabilities in commonly used software.
Their standard practice is to inform the vendor and work with them to help fix the vulnerability.
However, after a hard deadline of 90 days, the details of the vulnerability will be released to the public~\citep{zero2020policy}.
The effects are two-fold.
First, the deadline encourages faster patch development by the vendor.
Second, the techniques used can be studied to prevent similar vulnerabilities in the future.

\section{A note on licensing}

During the collection of our data set, we ran into an interesting question to which we could not find a satisfying answer.
We generated samples that are intrinsically designed to be as close as possible to the original data set.
So, when distributing these samples (or the models that generated them), it is unclear whether the original license still applies. The data is obviously not the original data.
Yet, it sounds remarkably like it.
To the best of our knowledge, this question has not been addressed by the machine learning or legal community.

\section{Training details}

We trained \gmms{} using gradient descent for ten epochs, with a batch size of 128, minimizing the negative log-likelihood of the data distribution.
We use 128 mixture components and learn the diagonal covariance matrix of each distribution. 
We double the number of components to 256 for the leave-one-out experiments to compensate for the more difficult task. 
When training \rawnet{} models we use the model configuration proposed by~\citet{tak2021rawnet2}.
We minimize the binary cross-entropy using gradient descent, a batch size of 128, and training for ten epochs.
During training we measure the validation accuracy over a hold-out set and restore the best performing model at the end of the training.
We use the Adam~\citep{kingma2014adam} optimizer with an initial learning rate of 0.001 when training \gmm{} models and 0.0001 when training \rawnet{}.
Additionally, we utilize weight decay (0.0001) when training \rawnet{}, following~\citet{tak2021rawnet2}.

We resample all audio files to 16kHz and remove silence parts that are longer than two seconds.
When converting the audio files to \mfcc{}/\lfcc{} features, we use the parameters proposed by~\citet{sahidullah2015comparison}.
We extract 20 \ac{LFCC}/\ac{MFCC} features and compute delta-/double-delta-features, \cf{} Section 2.
When training directly on raw audio, we also resample and remove silence from the audio.
Otherwise, we follow~\citet{tak2021rawnet2} and either pad or trim the data to 4s.

We trained all our models on a machine running Ubuntu 18.04.5 LTS, with a AMD Ryzen 7 3700X 8-Core Processor, a GeForce RTX 2080Ti, and 64GB of RAM.
The implementation of our models was performed in PyTorch 1.8.1, using the torchaudio extension in version 0.8.1~\cite{pytorch}.
Training a model for ten epochs on 10,000 audio samples takes roughly half an hour.
We do not implement the \rawnet{} models but instead utilize an open-source version provided by the authors~\citep{tak2021rawnet2}.
The code can be found online, and we do not redistribute it.

\begin{table}[t!]
    \centering
    \caption{
    \textbf{Equal Error Rate (EER) of the baseline classifier on different subset (MFCC).}
    We train a new GMM model for each training set and use the log-likelihood ratio to score every sample.
    For each data set we compute the EER, best possible result is 0.0, worst is 1.0, the lower the better.
    Additionally, we compute the average EER (aEER) over all sets.
    }
    \resizebox{\linewidth}{!}{\begin{tabular}{@{}lrrrrrrrrrrrr@{}}
        \toprule
 &  \multicolumn{8}{c}{\lj} & \phantom{a} &\multicolumn{2}{c}{\jsut}\\ 
\cmidrule{2-9} \cmidrule{11-12} \rule{0pt}{2.2ex}
Training Set & \multicolumn{1}{c}{\Mel{}} & \multicolumn{1}{c}{\Mel{} (L)} & \multicolumn{1}{c}{\FBMel{}} & \multicolumn{1}{c}{\MBMel{}} &   \multicolumn{1}{c}{\hifi{}} &\multicolumn{1}{c}{\wav{}} & \multicolumn{1}{c}{\PWG{}} & \multicolumn{1}{c}{\FTTS{}} & & \multicolumn{1}{c}{\MBMel{}} & \multicolumn{1}{c}{\PWG{}} & \multicolumn{1}{c}{\phantom{a}\textbf{aEER}} \\ \midrule 
\rule{0pt}{2.4ex}\Mel{} & \gray{0.332} & \textbf{0.309} & 0.476 & 0.439 & 0.458 & 0.513 & 0.388 & 0.143 &  & 0.077 & 0.074 & 0.341 \\
\rule{0pt}{2.4ex}\Mel{} (L) & \textbf{0.295} & \gray{0.177} & 0.437 & 0.440 & 0.447 & 0.515 & 0.358 & \textbf{0.092} &  & 0.146 & 0.176 & 0.332 \\
\rule{0pt}{2.4ex}\MBMel{} & 0.481 & 0.466 & \gray{0.025} & 0.371 & 0.318 & \textbf{0.069} & 0.144 & 0.346 &  & 0.184 & 0.259 & 0.257 \\
\rule{0pt}{2.4ex}\FBMel{} & 0.434 & 0.423 & 0.313 & \gray{0.270} & 0.351 & 0.324 & 0.281 & 0.340 &  & 0.405 & 0.434 & 0.360 \\
\rule{0pt}{2.4ex}\hifi{} & 0.468 & 0.458 & 0.313 & 0.386 & \gray{0.252} & 0.288 & 0.256 & 0.285 &  & 0.225 & 0.253 & 0.322 \\
\rule{0pt}{2.4ex}\PWG{} & 0.503 & 0.508 & \textbf{0.092} & 0.417 & 0.359 & \gray{0.014} & \textbf{0.190} & 0.427 &  & \textbf{0.035} & \textbf{0.053} & 0.241 \\
\rule{0pt}{2.4ex}\wav{} & 0.437 & 0.421 & 0.120 & \textbf{0.334} & \textbf{0.277} & 0.112 & \gray{0.053} & 0.194 &  & 0.067 & 0.105 & \textbf{0.214} \\

        \bottomrule
    \end{tabular}}
    \begin{flushleft}
    \scriptsize{When the distribution is part of the training set we highlight it in \gray{gray}. For other results, we highlight the best results per column in \textbf{bold}.}
    \end{flushleft}
    \label{tab:mfcc_results}
\end{table}
 
\section{\mfcc{} results}
Since \mfcc{} features are commonly used for, e.g., automatic speech recognition, we also evaluated them.
However, we found them to be strictly outperformed by \lfcc{} features.
The results are display in Table~\ref{tab:mfcc_results}.
When comparing the overall performance, \ie, the lowest average \ac{EER} (aEER), we can observe that \PWG{} ($0.241$), \MBMel{} ($0.257$), and, \wav{} ($0.214$) serve as the best priors for the entire data set.
However, they all perform significantly worse on the \Mel{}, the \Mel{} (L) data sets.
This trend is reversed for \Mel{} and \Mel{} (L), where they achieve the best results on each other ($0.295$ and $0.309$, respectively) and dropping performance on other data sets ($\sim 0.400$; up to $0.515$ on \wav{}).
\FBMel{} does not perform particularly well on any data set.

The similarities between \PWG{} and \wav{} are intuitive.
The \wav{} architecture is heavily inspired by WaveNet (the generator network of \PWG{}).
Yet, the best results for both \PWG{} ($0.092$) and \wav{} ($0.069$) are obtained by the models trained on \MBMel{} and \FBMel{}.
We hypothesize that the auxiliary loss forces them to generate samples more in line with \wav{} and \PWG{}.
Surprisingly neither \FBMel{} nor \MBMel{}, generalize to the \Mel{} (L) data or \MBMel{} data sets, despite using similar generator architectures.

\section{Phone simulation results}
\begin{table}[t!]
    \centering
    \caption{
        \textbf{Equal Error Rate (EER) for the phone recording simulation (\lfcc{}).}
        We use the models from the out-of-distribution experiments.
    }
    \resizebox{\linewidth}{!}{\begin{tabular}{@{}lrrrrrrr@{}}
        \toprule
Test Set & \multicolumn{1}{r}{\Mel{}} & \multicolumn{1}{r}{\Mel{} (L)} & \multicolumn{1}{r}{\FBMel{}} & \multicolumn{1}{r}{\MBMel{}} &\multicolumn{1}{r}{\hifi{}} &   \multicolumn{1}{r}{\wav{}} & \multicolumn{1}{r}{\PWG{}} \\ \midrule
\FTTS{} &       0.000 &    0.000 &     0.001 & 0.000 & 0.006  & 0.000 & 0.000 \\
\rule{0pt}{2.4ex}\jsut{} \MBMel{} &   0.001 & 0.002 &  0.003 & 0.001 & 0.003 & 0.001 & 0.000  \\
\rule{0pt}{2.4ex}\jsut{} \PWG{} & 0.003 & 0.002 &  0.002 & 0.002 & 0.003 &  0.003 & 0.001  \\
        \bottomrule
    \end{tabular}}
    \begin{flushleft}
    \scriptsize{The columns represent the left-out-set during training}
    \end{flushleft}
    \label{tab:phone}
    \vspace{-1em}
\end{table}

\begin{table}[t!]
    \centering
    \caption{
        \textbf{Equal Error Rate (EER) for the phone recording simulation (\rawnet{}).}
        We use the models from the out-of-distribution experiments.
    }
    \resizebox{\linewidth}{!}{\begin{tabular}{@{}lrrrrrrr@{}}
        \toprule
Test Set & \multicolumn{1}{r}{\Mel{}} & \multicolumn{1}{r}{\Mel{} (L)} & \multicolumn{1}{r}{\FBMel{}} & \multicolumn{1}{r}{\MBMel{}} &\multicolumn{1}{r}{\hifi{}} &   \multicolumn{1}{r}{\wav{}} & \multicolumn{1}{r}{\PWG{}} \\ \midrule
\FTTS{} &       0.144 &    0.549 &     0.357 & 0.201 & 0.180 & 0.330 & 0.201 \\
\rule{0pt}{2.4ex}\jsut{} \MBMel{} &   0.065 & 0.898 &  0.842 & 0.028  & 0.915 & 0.911  & 0.028 \\
\rule{0pt}{2.4ex}\jsut{} \PWG{} & 0.159 & 0.835 &  0.740 & 0.008 & 0.937 & 0.932  & 0.008 \\
        \bottomrule
    \end{tabular}}
    \begin{flushleft}
    \scriptsize{The columns represent the left-out-set during training}
    \end{flushleft}
    \label{tab:phone_rawnet}
    \vspace{-1em}
\end{table} 
Table~\ref{tab:phone} presents the results of the phone simulation experiment.
We evaluate the models from the out-of-distribution evaluation.
The columns represent the left-out-set for the corresponding model and measure the performance on our three test sets.

\section{Spectrograms}

Here we plot the spectrograms of an audio file (\lj{} \datasetstyling{008-0217}) for the training data and the different generative networks.
Notice the differences, especially in the higher frequencies and the horizontal artifacts produced by \Mel{} and \wav{}.

\newcommand{\specfigsize}{.95\linewidth}

\begin{figure}[h!tp]
  \centering
  \includegraphics[width=\specfigsize]{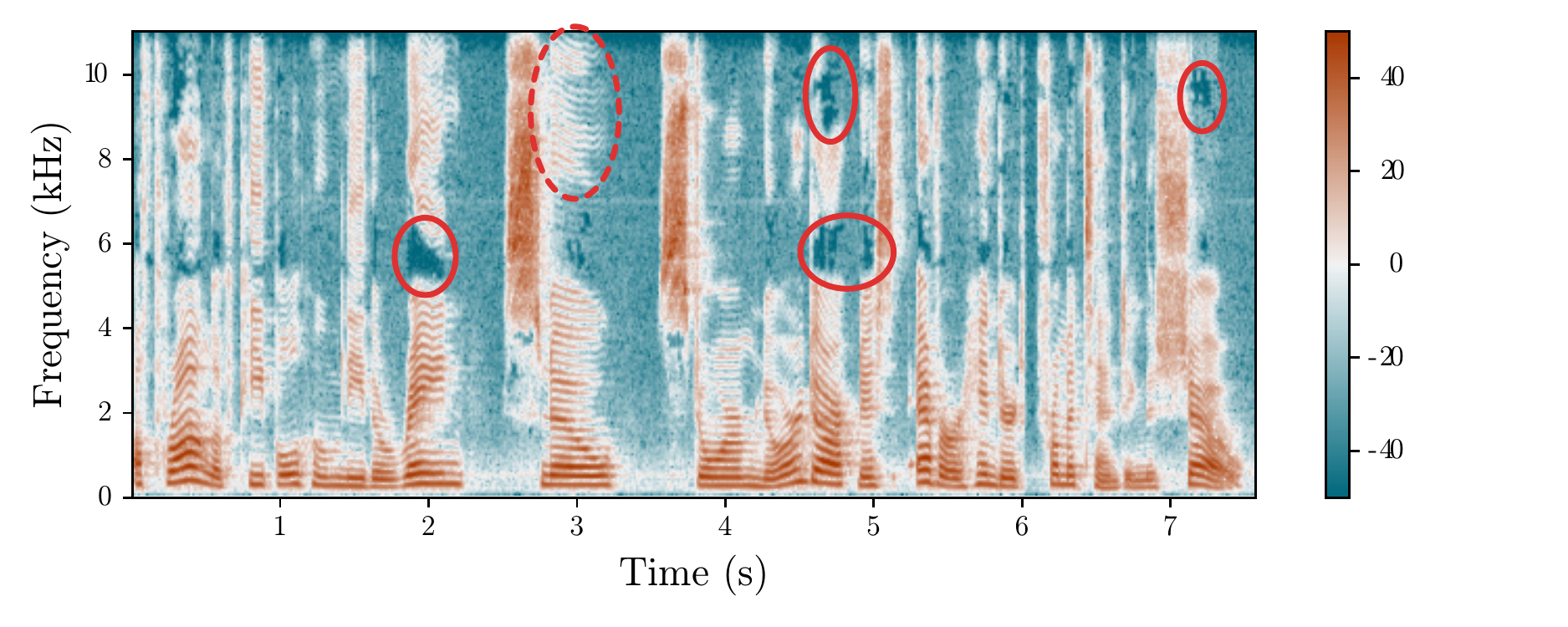}
  \caption{Original}
\end{figure}

\begin{figure}[h!tp]
  \centering
  \includegraphics[width=\specfigsize]{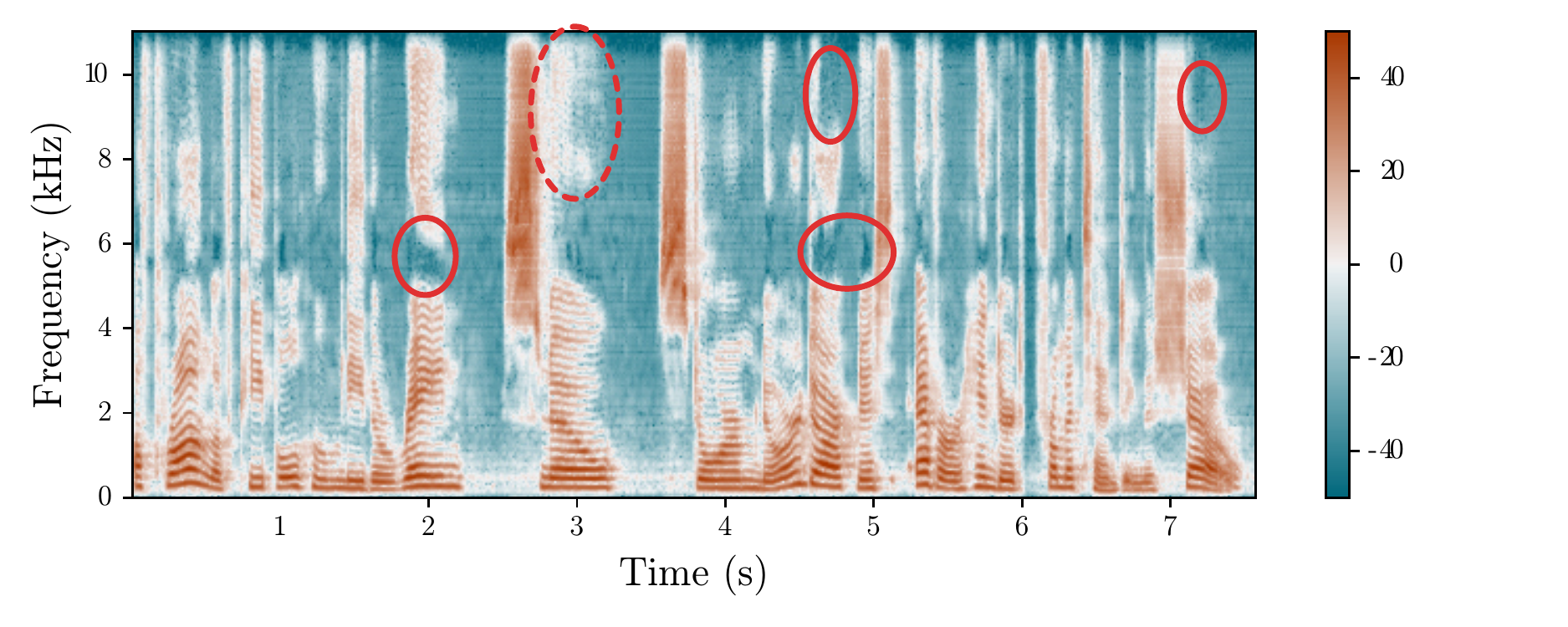}
  \caption{\Mel{}}
\end{figure}

\begin{figure}[h!tp]
  \centering
  \includegraphics[width=\specfigsize]{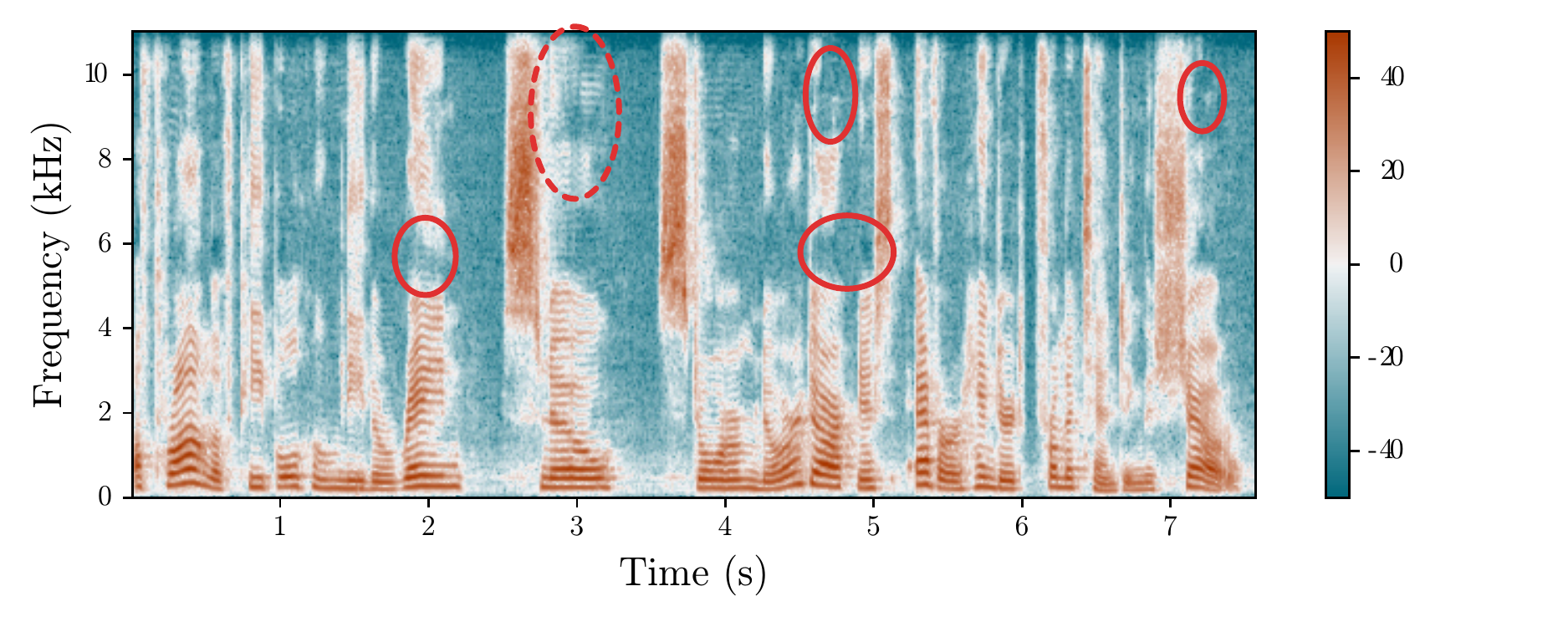}
  \caption{\FBMel{}}
\end{figure}
\hfill
\begin{figure}[h!tp]
  \centering
  \includegraphics[width=\specfigsize]{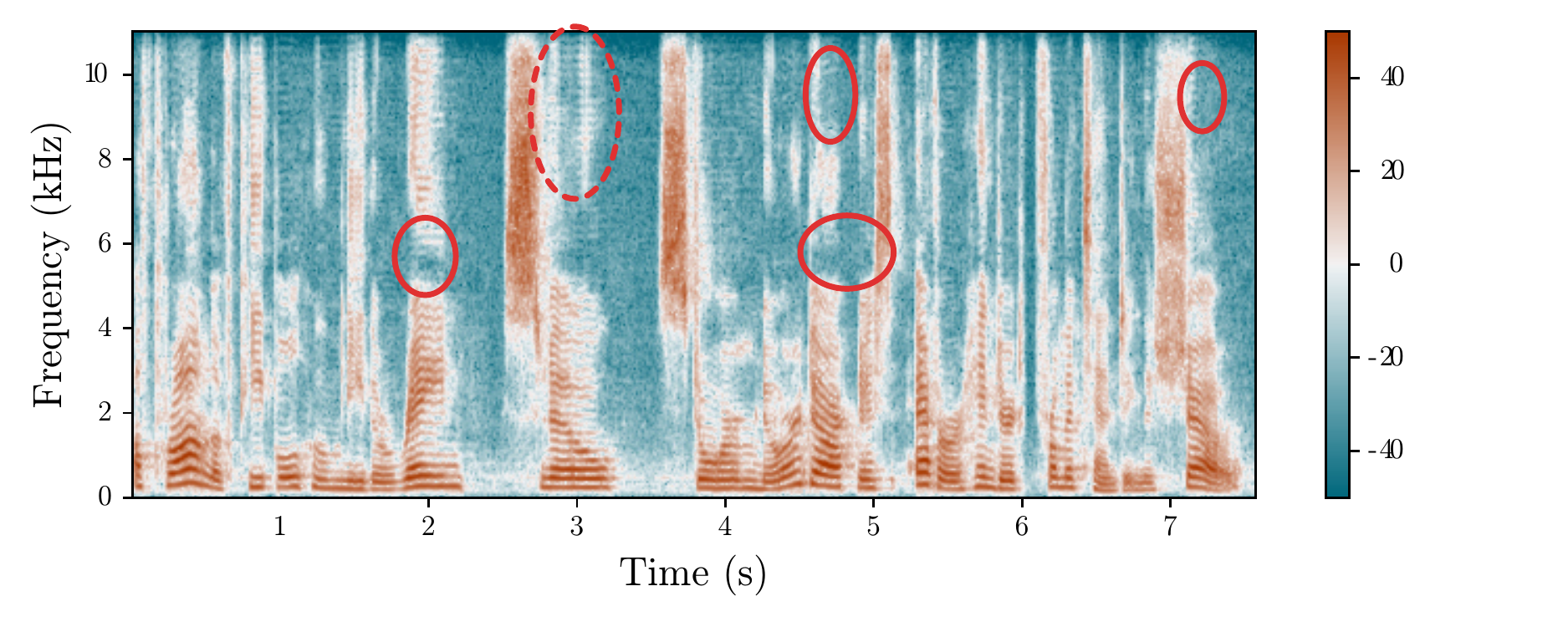}
  \caption{\MBMel{}}
\end{figure}

\bigskip

\begin{figure}[h!tp]
\centering
\includegraphics[width=\specfigsize]{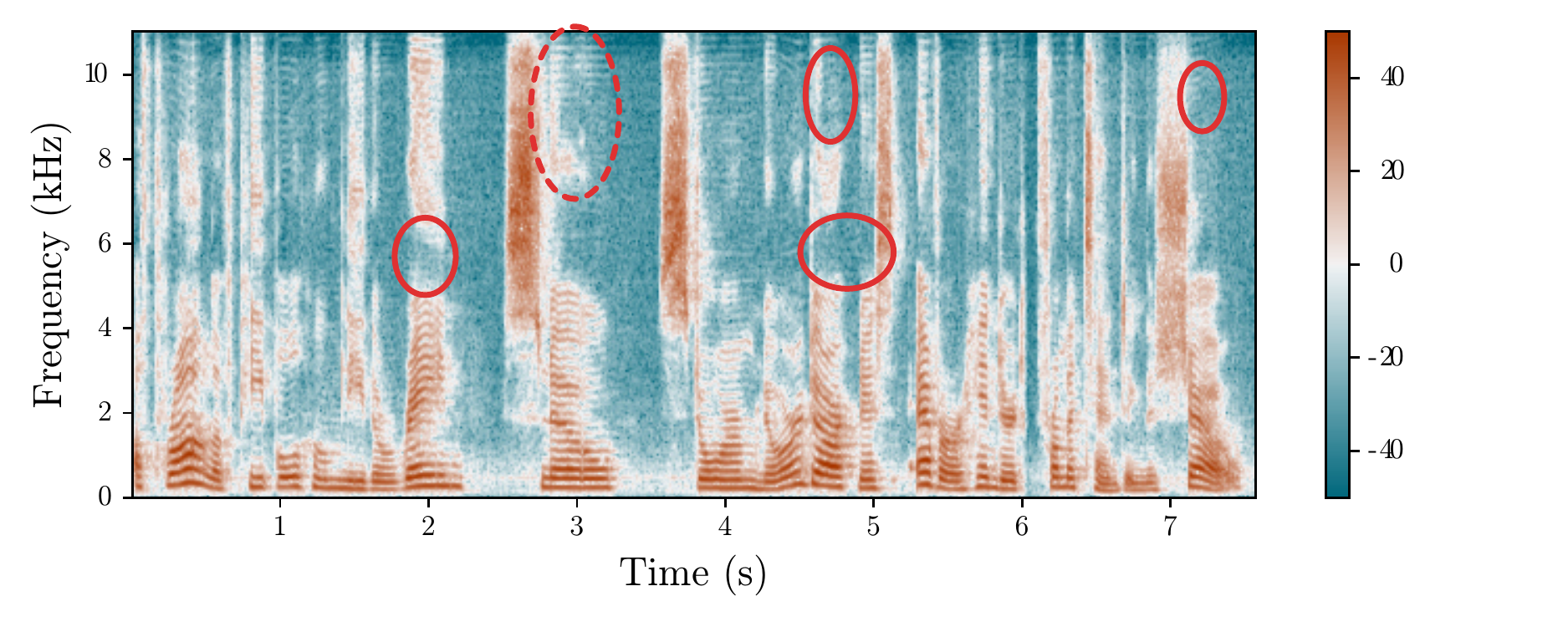}
\caption{\hifi{}}
\end{figure}
\hfill
\begin{figure}[h!tp]
  \centering
  \includegraphics[width=\specfigsize]{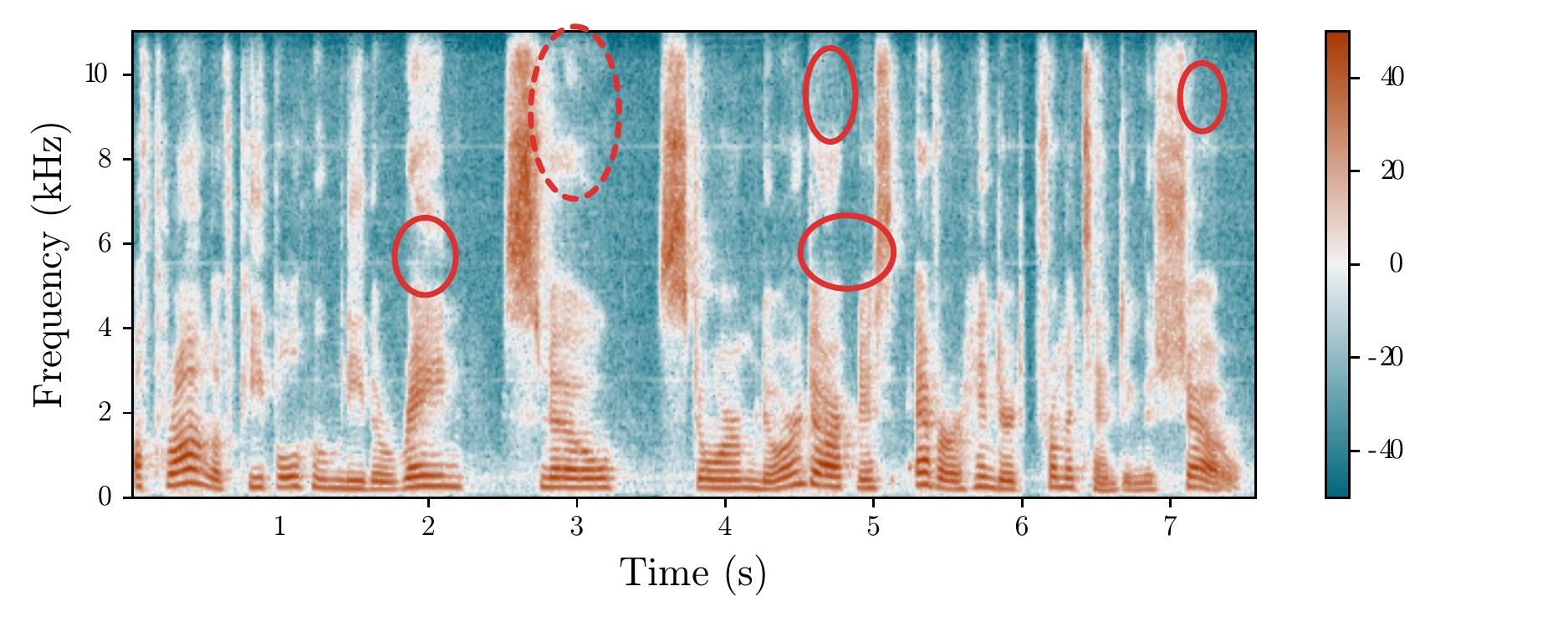}
  \caption{\wav{}}
\end{figure}

\bigskip

\begin{figure}[h!tp]
\centering
\includegraphics[width=\specfigsize]{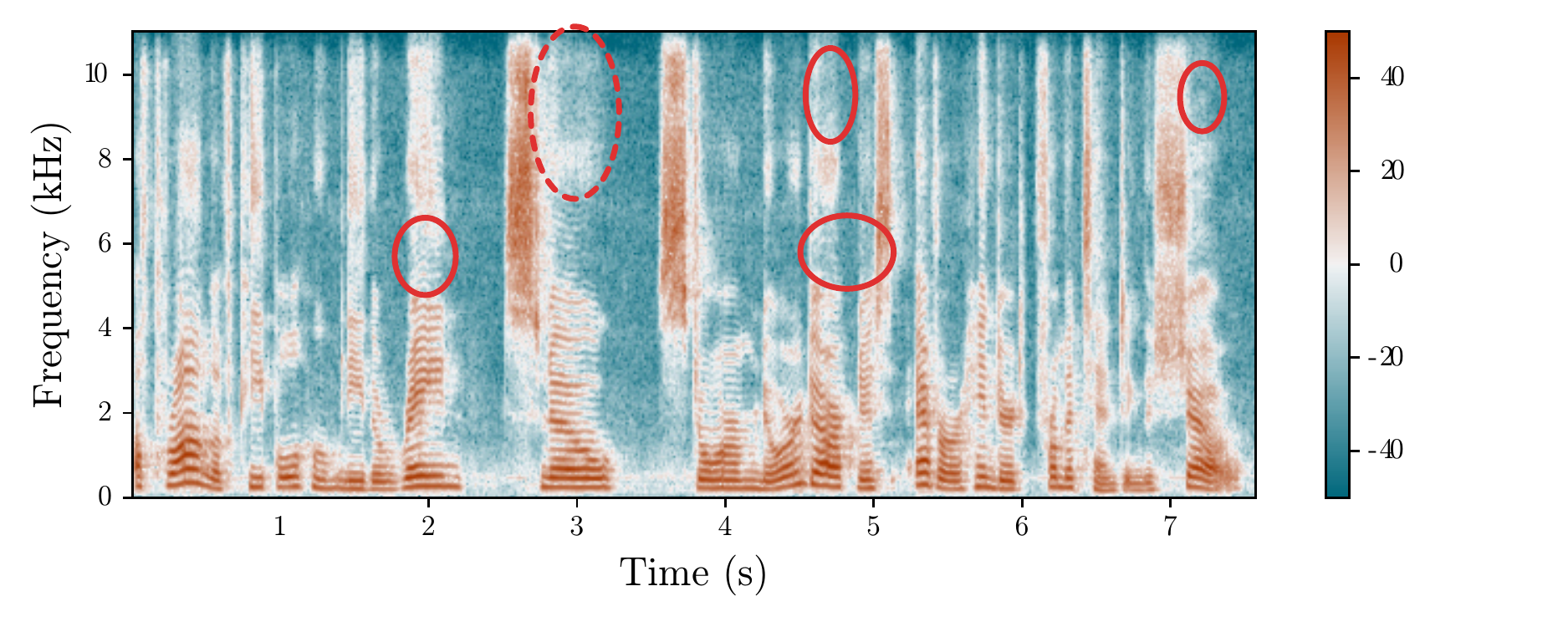}
\caption{\PWG{}}
\end{figure}

 \begin{figure}[t!]
  \centering
\newcommand{\histfigsize}{.89\columnwidth}
  \begin{subfigure}{\histfigsize}
      \centering
      \includegraphics[width=.99\linewidth]{content/figures/media/histograms/melgan_differnce.pdf}
  \end{subfigure}
  \begin{subfigure}{\histfigsize}
      \centering
      \includegraphics[width=.99\linewidth]{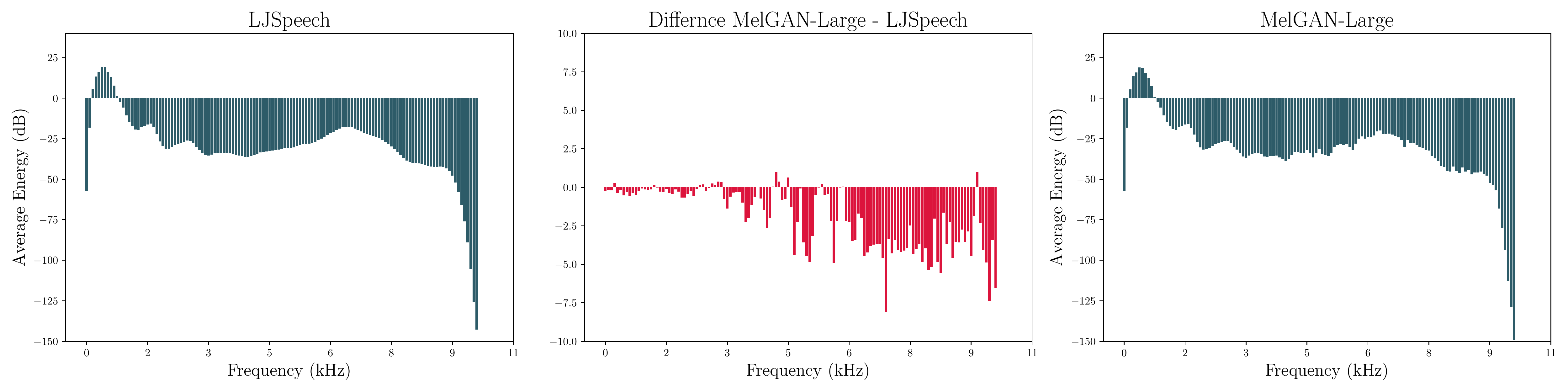}
  \end{subfigure}
  \begin{subfigure}{\histfigsize}
      \centering
      \includegraphics[width=.99\linewidth]{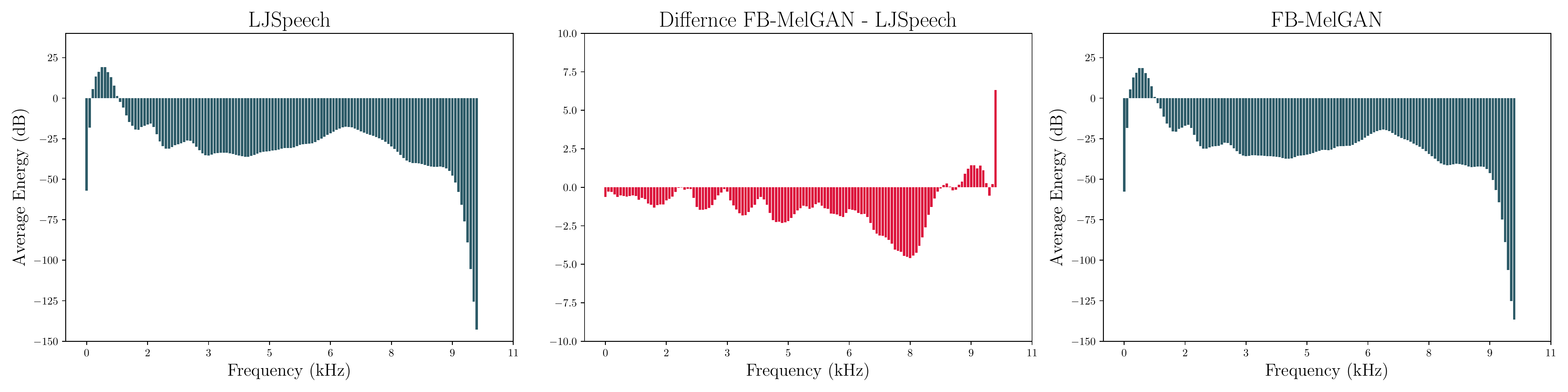}
  \end{subfigure}
  \begin{subfigure}{\histfigsize}
      \centering
      \includegraphics[width=.99\linewidth]{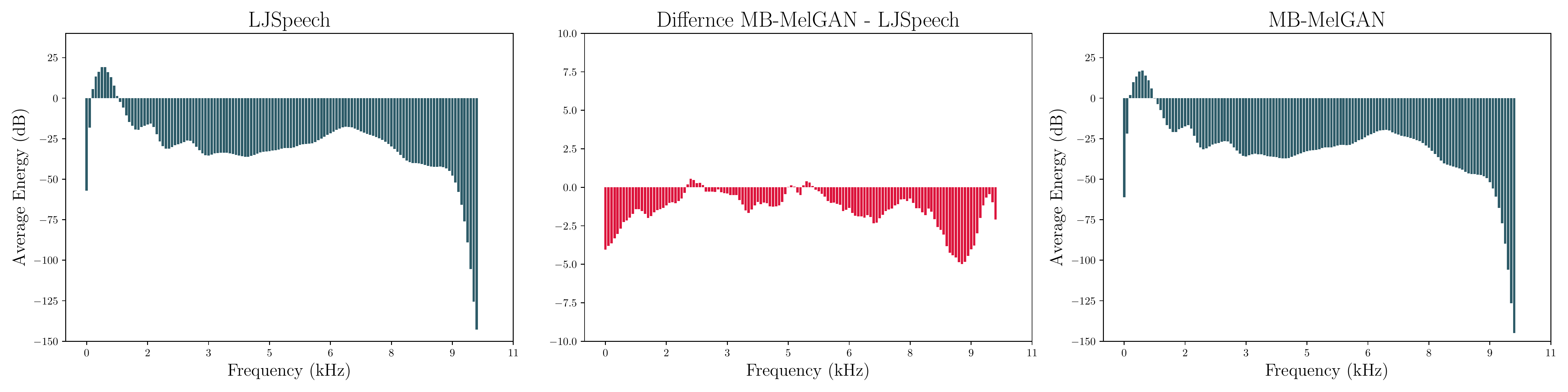}
  \end{subfigure}
  \begin{subfigure}{\histfigsize}
      \centering
      \includegraphics[width=.99\linewidth]{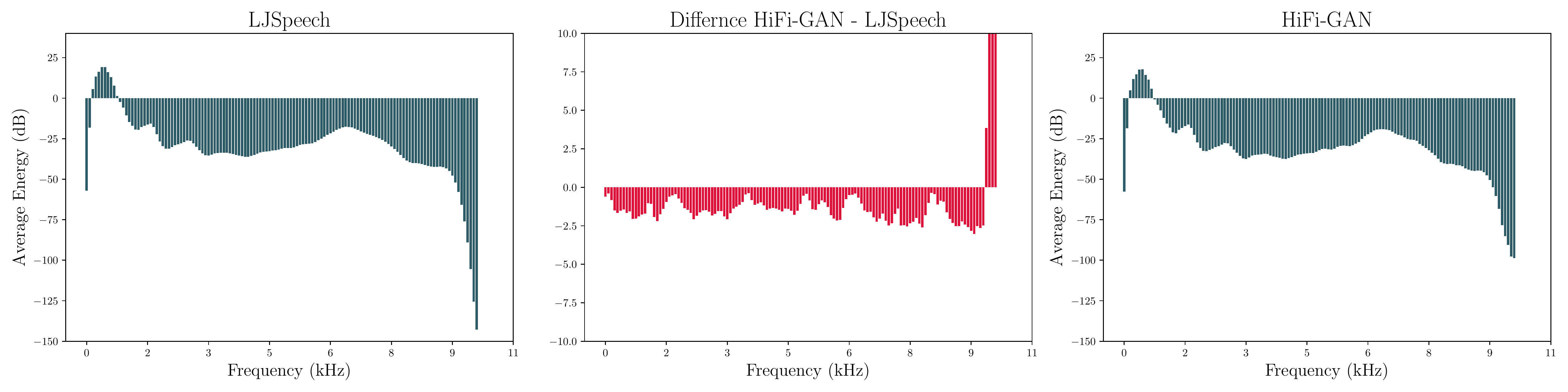}
  \end{subfigure}
  \begin{subfigure}{\histfigsize}
      \centering
      \includegraphics[width=.99\linewidth]{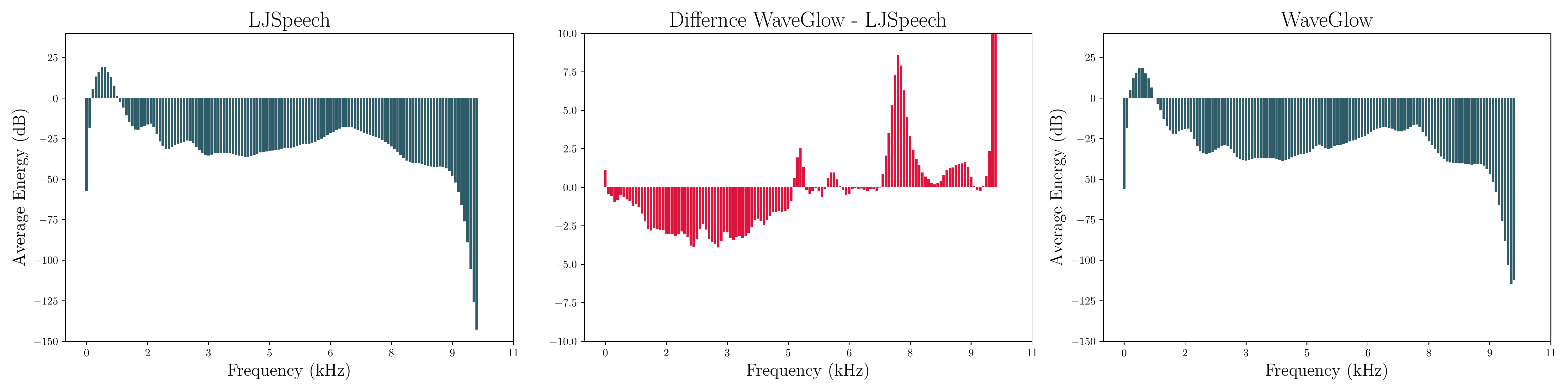}
  \end{subfigure}
\begin{subfigure}{\histfigsize}
      \centering
      \includegraphics[width=.99\linewidth]{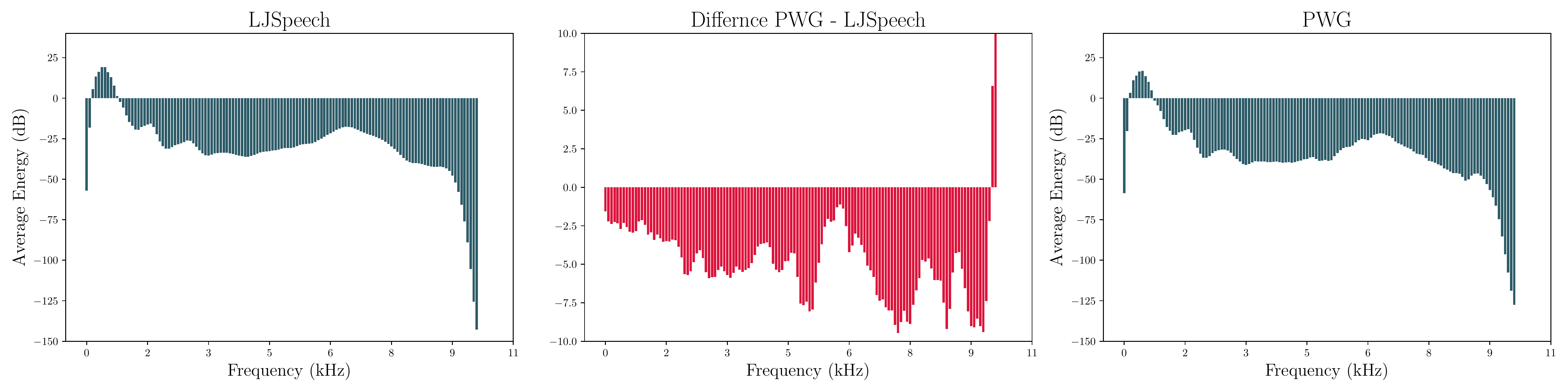}
  \end{subfigure}
\caption{\textbf{Average energy per frequency bin.} 
  We show the average energy per frequency bin in dB.
  Additionally, we plot the difference to the original data (\lj{}).
  }
\end{figure} 
\clearpage
\section{Attribution}

These are the full-size version of the attribution plots used in Section 4.3.
Note the spread out the attention of the \Mel{} classifier, the transition to narrow band attribution, and the balance of the classifier trained on \FBMel{}.

\begin{figure}[h!]
  \centering
\begin{subfigure}{\columnwidth}
      \centering
      \includegraphics[width=.75\linewidth]{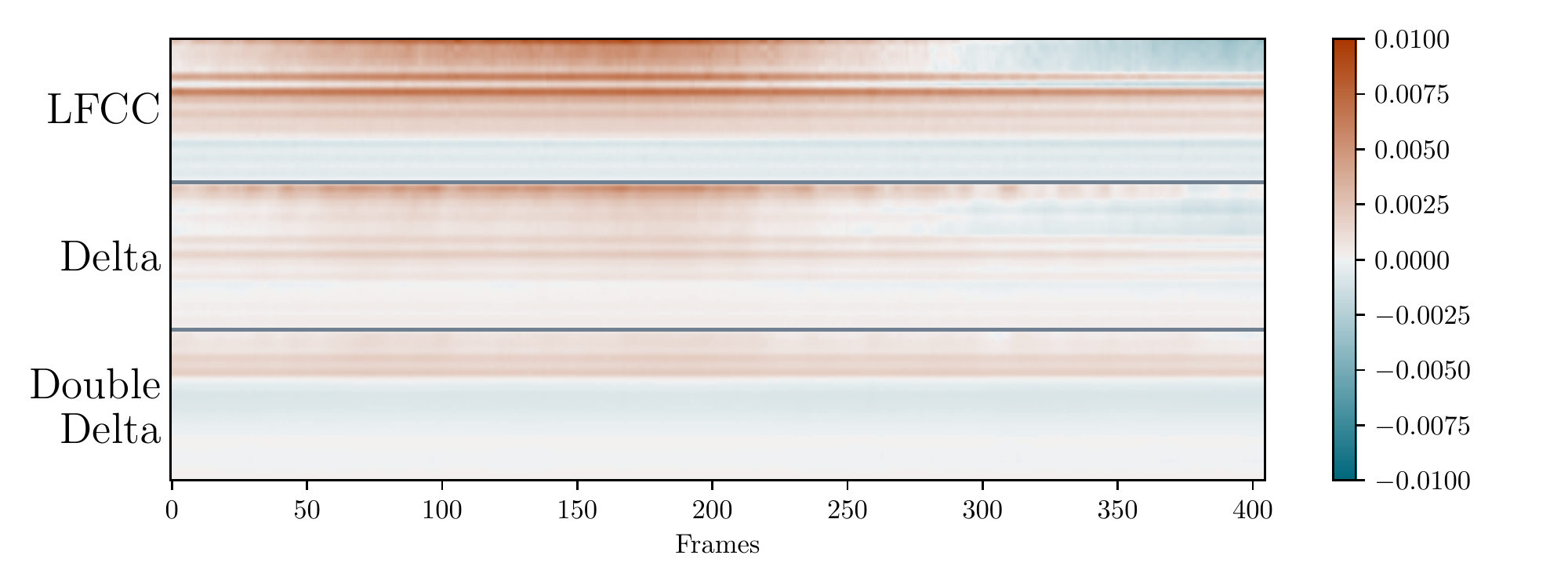}
      \caption{\Mel{} (L)}
  \end{subfigure}
\begin{subfigure}{\columnwidth}
      \centering
      \includegraphics[width=.75\linewidth]{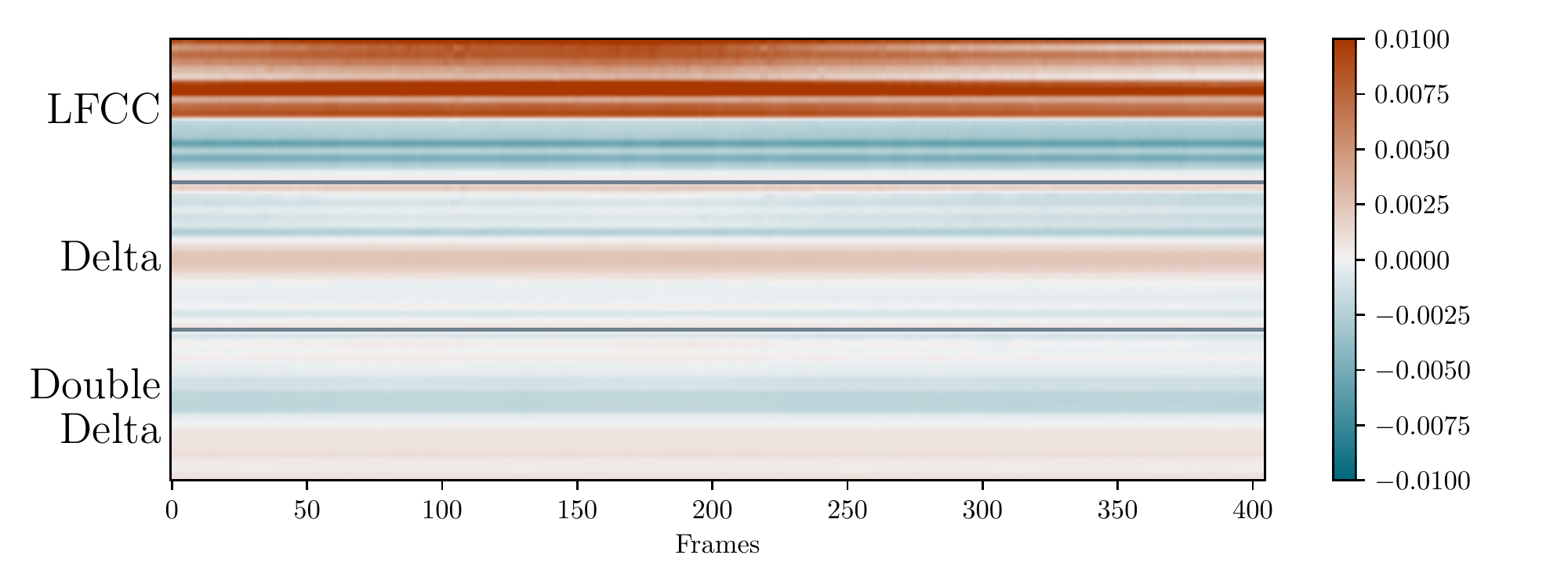}
      \caption{\FBMel{}}
  \end{subfigure}
\hfill
\begin{subfigure}{\columnwidth}
      \centering
      \includegraphics[width=.75\linewidth]{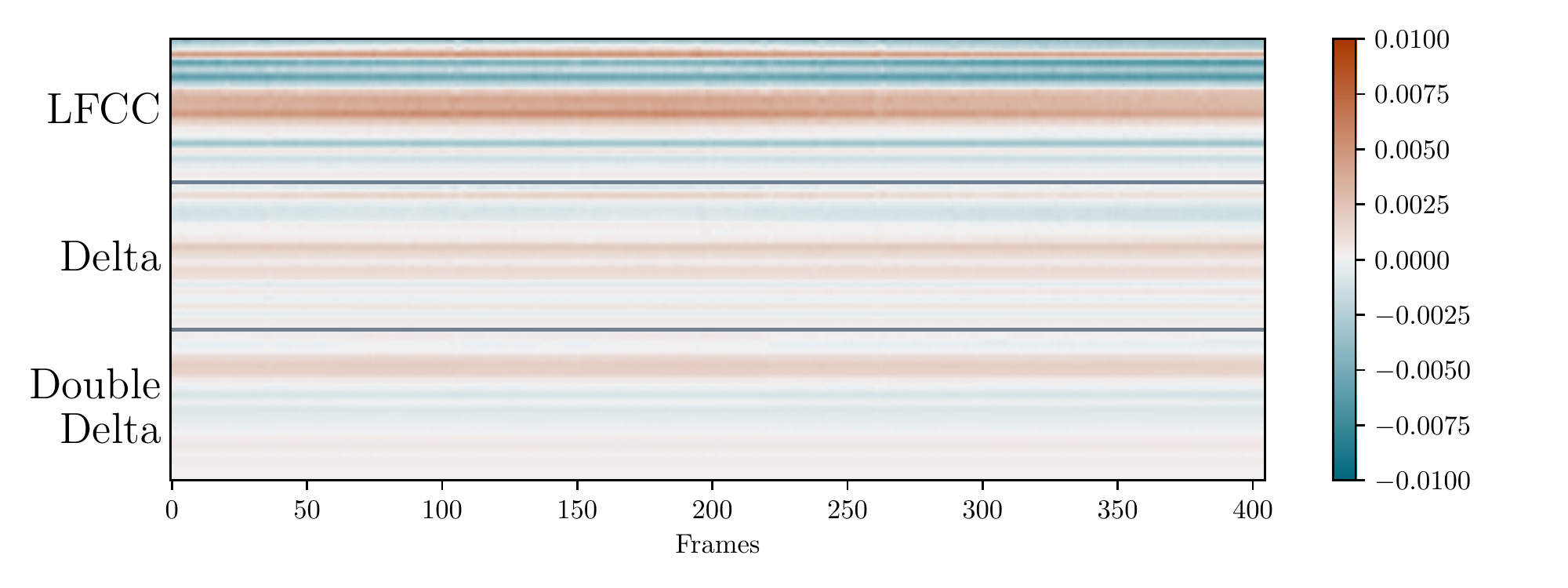}
      \caption{\MBMel{}}
  \end{subfigure}
\begin{subfigure}{\columnwidth}
      \centering
      \includegraphics[width=.75\linewidth]{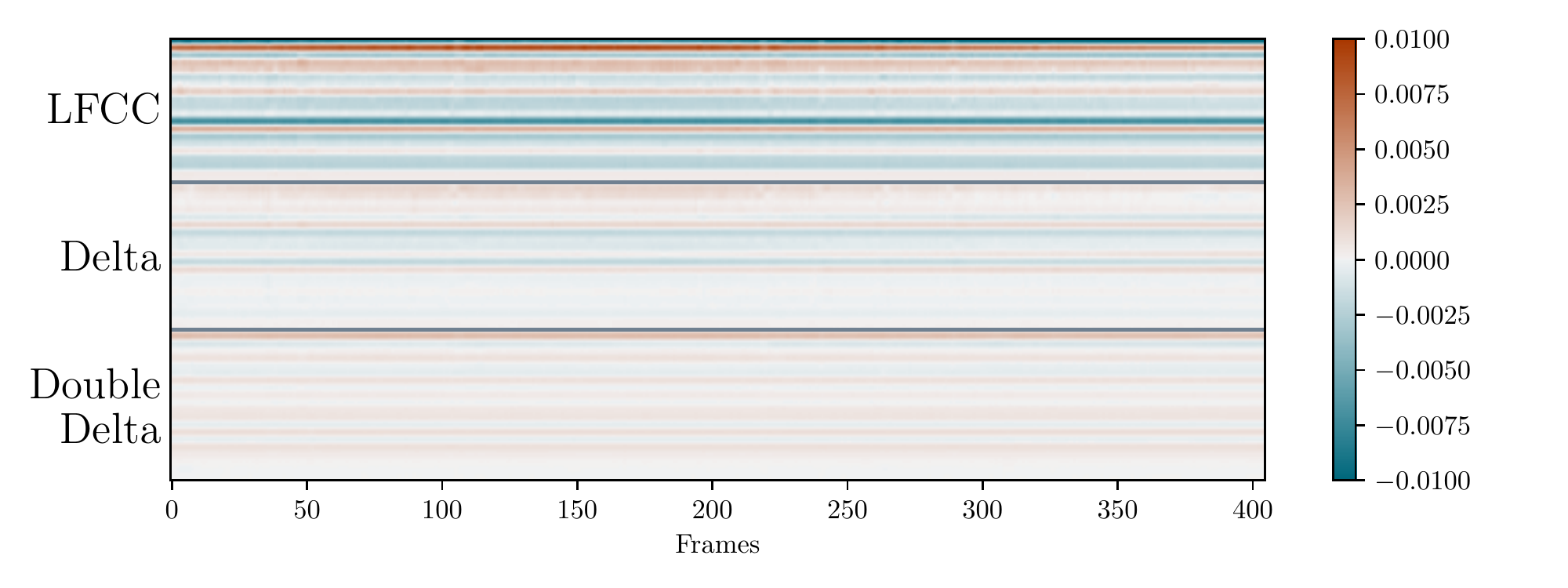}
      \caption{\PWG{}}
  \end{subfigure}
\end{figure}

\clearpage

\section{Filterbanks}

Here we show a visual representation of the triangular filterbanks used to compute the \mfcc{} and \lfcc{} features.

\begin{figure}[h]
    \centering
    \includegraphics[width=.8\linewidth]{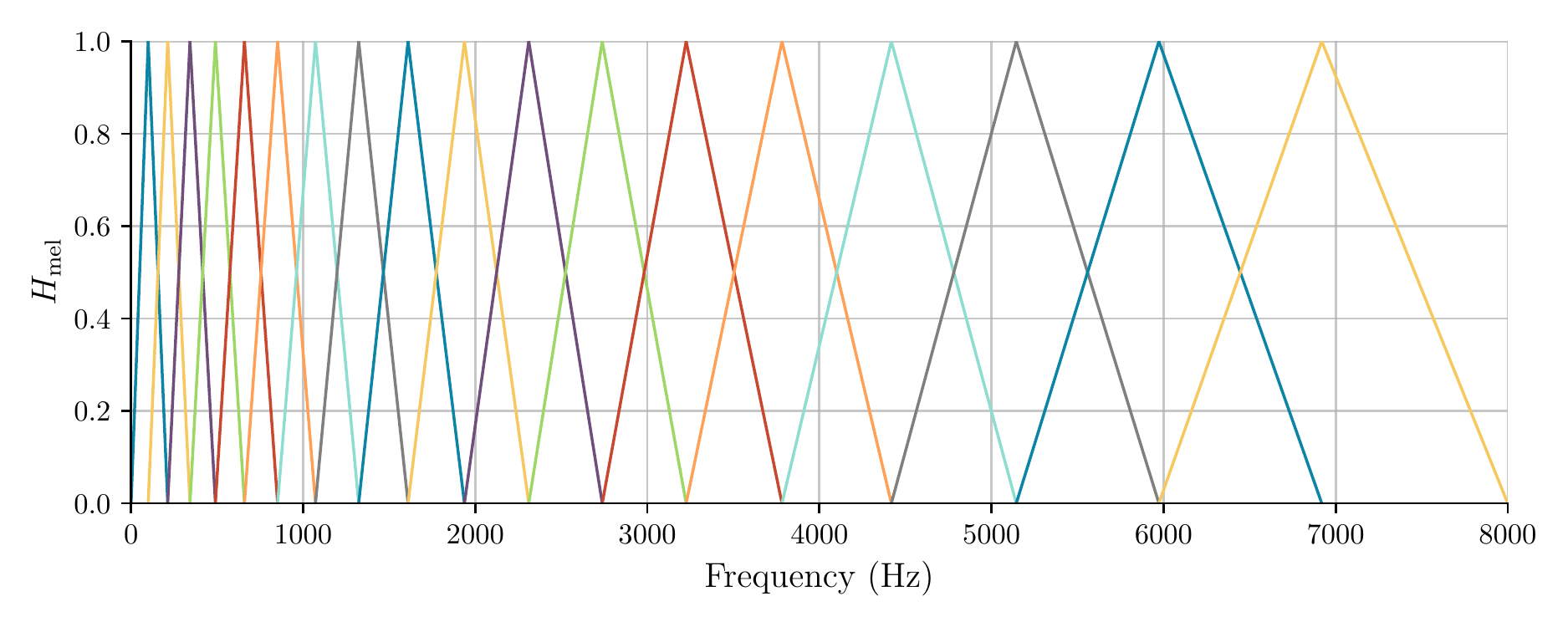}
    \caption{Mel filterbank}
\end{figure}

\begin{figure}[h]
    \centering
    \includegraphics[width=.8\linewidth]{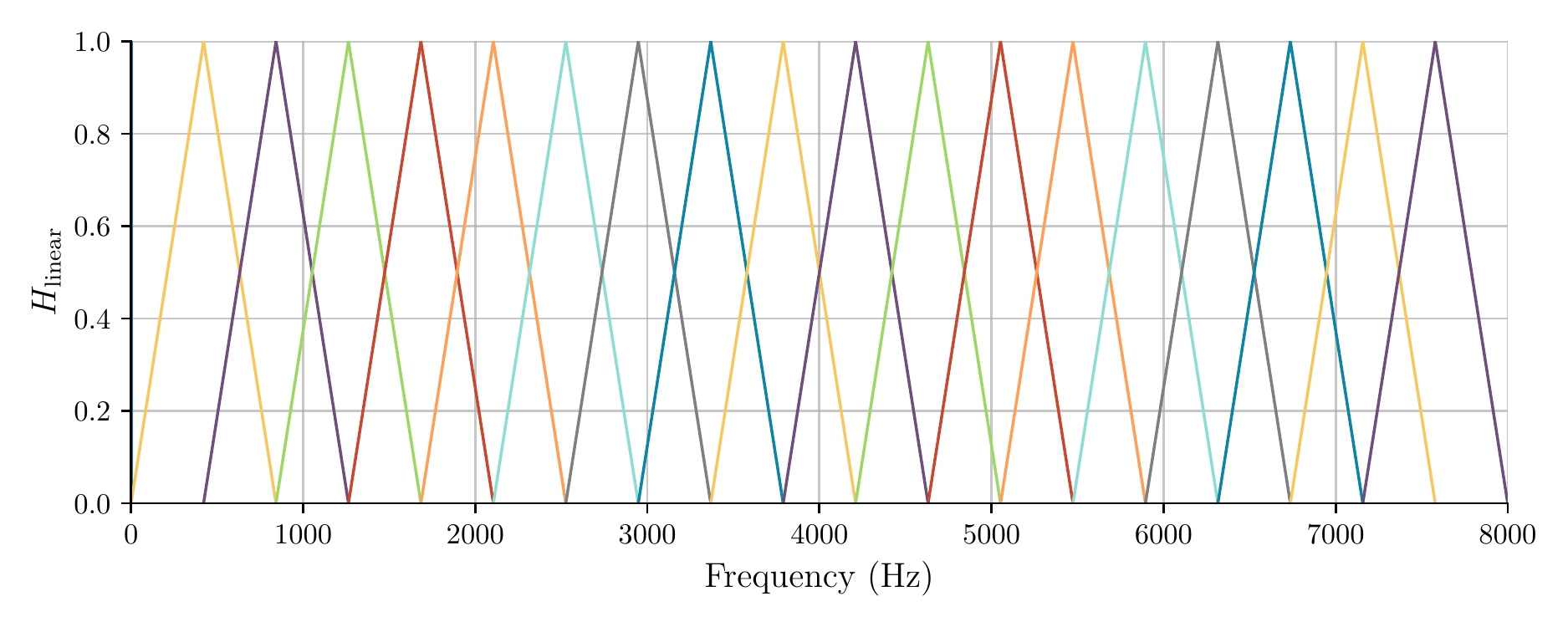}
    \caption{Linear filterbank}
\end{figure}

\end{document}